\documentclass[]{fairmeta}
\usepackage[table]{xcolor}
\usepackage{longtable}
\usepackage{tikz}
\usetikzlibrary{calc, positioning}

\usepackage{booktabs}
\usepackage{multirow}
\usepackage{makecell}
\usepackage{url}
\usetikzlibrary{shapes.geometric}   
\definecolor{darkgreen}{RGB}{0, 100, 0}
\definecolor{lightyellow}{RGB}{255, 255, 204}

\usepackage{pgfplots}
\pgfplotsset{compat=1.18}

\definecolor{colorDirect}{RGB}{44, 82, 130}      
\definecolor{colorTruncated}{RGB}{221, 107, 32}  
\definecolor{colorFull}{RGB}{47, 133, 90}        

\definecolor{colorWithout}{HTML}{4C72B0}
\definecolor{colorWith}{HTML}{55A868}

\usepackage{graphicx}

\usepackage{hyperref}
\usepgfplotslibrary{groupplots}

\definecolor{c700M}{HTML}{1f77b4}  
\definecolor{c14B}{HTML}{2ca02c}   
\definecolor{c23B}{HTML}{ff7f0e}   
\definecolor{c37B}{HTML}{9467bd}   
\definecolor{c55B}{HTML}{d62728}   
\definecolor{c9B}{HTML}{8c564b}    

\title{MetaEncoder: Exploring the Limit of Bi-Encoders for Multimodal System One Decision Making with Natural Language Interface}

\author[]{Jianpeng Cheng} 
\author[]{Guangyu Sun}
\author[]{Aashu Singh}
\author[]{Benyu Zhang} 
\author[]{Haixing Dai}
\author[]{Hossein Mansour}
\author[]{Jiangfan Zhang}
\author[]{Shlok Kumar Mishra}
\author[]{Wei Sun} 
\author[]{Xuanming Cui}  
\author[]{Yanli Liu}  
\author[]{Qi Guo}
\author[]{Max Xiangjun Fan} 
\author[]{Jun Xiao}

\affiliation[1]{Meta AI}

\abstract{
System One models prioritize fast, low-latency decision making and calibrated probability distributions over token generation. While prevailing systems rely on structured schemas and typed outputs, the trade-offs among candidate-scoring mechanisms remain poorly understood. In this work, we systematically investigate the scaling dynamics, accuracy, and inference efficiency of three scoring architectures adapted from a pre-trained decoder: bi-encoders, cross-encoders, and constrained decoders. Our experiments show clear performance scaling with model size across all three architectures. Accuracy follows clear trade-offs too: cross-encoders consistently achieve the highest accuracy across all model sizes, while bi-encoders and constrained decoders perform similarly as we scale up training tokens. Efficiency-wise, bi-encoders maintain near-constant inference latency as candidate pools grow because candidate embeddings can be cached offline, whereas the inference costs of cross-encoders and constrained decoders both scale quadratically with candidate size. The efficiency advantage makes bi-encoders uniquely suited to bridge small closed-set decisions and massive open-set retrieval. For closed-set tasks, we find it critical to include all candidate options in the request prompt to make bi-encoders competitive. Guided by these insights, we built \textbf{MetaEncoder}, a multimodal encoder fine-tuned from a \texttt{Muse-Glimmer} decoder via unidirectional contrastive learning to match request and candidates. We introduce packed contrastive training to maximize the training efficiency of dynamically allocated negative candidates, enabling memory-efficient full fine-tuning of a 30B decoder into an encoder. We thoroughly evaluate MetaEncoder on 12 benchmark suites spanning closed-set decision-making, multimodal understanding, and open-set retrieval. MetaEncoder outperforms state-of-the-art multimodal embedding baselines on decision making tasks as well as MMEB-v3 image and video retrieval, while scaling to open candidate spaces which are far too large for current decision models to handle. 
}

\date{October 9, 2026}
\huggingface{\url{https://huggingface.co/facebook/meta-encoder}}
\github{\url{https://github.com/facebook/meta-encoder-eval}}

\begin{document}

\maketitle
\section{Introduction}
\label{sec:intro}

There is growing momentum around repurposing general-purpose candidate scoring models as \textbf{System One} models for rapid decision-making~\citep{TypeSafeAI2026a}. Rather than generating unconstrained text, a System One model maps an input directly to a constrained categorical choice and a calibrated probability distribution over candidate options. This paradigm matches the requirements of high-throughput production applications, such as agent routing, tool selection, policy-based content moderation, item ranking and action selection in embodied environments. For these latency-critical tasks, obtaining a reliable, calibrated score distribution from a single forward pass is often faster, more deterministic, and more robust than parsing unconstrained text generation outputs. 
However, candidate-scoring is not a new topic and its architectures generally fall into three distinct paradigms: \begin{enumerate} \item \textbf{Bi-encoders}~\citep{reimers2019sbert,karpukhin2020dpr,wang2024e5mistral} embed the request and each candidate independently into a shared embedding space, computing matching scores via dot-product or cosine similarity. In multimodal domains, vision-language embedders~\citep{jiang2024e5v,jiang2024vlm2vec,zhang2024gme} frame classification, visual question answering, and grounding as instruction-following retrieval over candidate representations. \item \textbf{Cross-encoders}~\citep{nogueira2019monobert,nogueira2020monot5} concatenate the request with each individual candidate and pass them through full attention layers. The score can be read-off from the binary logit representing the likelihood of the candidate. The approach trades substantial computational overhead for maximal representational interaction between a request and each option, serving as the de facto standard for high-precision reranking. \item \textbf{Constrained decoders}~\citep{hendrycks2021mmlu,robinson2023mcsb,reddy2024first} present all candidate options within a single prompt under discrete identifiers (e.g., A, B, C, \ldots) and read off the next-token probability distribution over those identifiers. Although constrained decoders allow joint attention over all options simultaneously, candidate capacity is fundamentally bounded by the context window of the backbone model. \end{enumerate}

For System One tasks, choosing an optimal scoring interface requires balancing accuracy, parameter scaling, and inference cost. However, a direct controlled comparison among bi-encoders, cross-encoders, and constrained decoders across these axes has been lacking. In this work, we conduct a systematic study evaluating these three scoring paradigms across model sizes, number of training tokens and candidate set cardinalities, uncovering the following central findings: 
\begin{itemize} 
\item \textbf{Model size and training token scaling:} Across all three scoring mechanisms, performance scales consistently with model size. With respect to training tokens, fine-tuning the scorer from a pre-trained decoder yields an immediate, sharp increase in accuracy, followed by more gradual gains as the number of training tokens increases.
\item \textbf{Scoring mechanisms:} Cross-encoders consistently deliver the highest accuracy across all evaluated model sizes. For the other two, bi-encoders and constrained decoders perform similarly at all model sizes as we scale up training tokens. 
\item \textbf{Inference cost:} The inference latency of cross-encoders and constrained decoders grows quadratically with the number of candidates, creating an inference bottleneck in candidate-rich environments. In contrast, bi-encoders maintain a near-constant inference cost when the number of candidates increases, since the candidate representations can be pre-computed and cached offline. 
\item \textbf{Decision making with bi-encoders:} While bi-encoders compress information into a single vector bottleneck, we demonstrate that their closed-set performance can match  constrained decoding when candidate options are explicitly prepended into the request prompt. This injects option-aware attention into the request representation. Crucially, while cross-encoders and constrained decoders are structurally restricted to bounded closed sets, the bi-encoder retains the flexibility to scale to massive open-set candidate pools.
\end{itemize}
Building on these insights, we train an open source \textbf{MetaEncoder}, a general-purpose multimodal encoder trained for instruction-following decision-making, classification and open retrieval. MetaEncoder fine-tunes a pre-trained \texttt{Muse-Glimmer}~\citep{MetaSuperintelligenceLab2026MuseGlimmer} decoder into an encoder using unidirectional contrastive learning to match request and candidates. To bridge closed-set and open-set scenarios within a single framework, MetaEncoder adapts its negative sampling strategy during training to differentiable closed-set and open-set problems. At inference time, the same model scales seamlessly from small closed sets to open corpora with massive candidates based entirely on prompt construction, handling multimodal queries and candidates alike.
We evaluate MetaEncoder across 12 benchmark suites. These span closed-set multimodal decision-making and understanding (JevBench~\citep{standhartinger2026jevbench}, MMLU~\citep{hendrycks2021mmlu} and MMLU-Pro~\citep{wang2024mmlu}, MMMU~\citep{yue2024mmmu} and MMMU-Pro~\citep{yue2025mmmu}, VideoMMMU~\citep{hu2025videommmu}, NaturalBench~\citep{li2024naturalbench}, and TempCompass~\citep{liu2024tempcompass}) and open-corpus retrieval (NanoBEIR~\citep{nanobeir2025}, MMEB-v3 image, video, and visdoc~\citep{huang2026mmeb}). MetaEncoder outperforms SOTA multimodal embedders~\citep{zhou2026wemm,li2026qwen3} across decision-making benchmarks, while achieving SOTA results on MMEB-V3 image and video retrieval—a capability beyond the scope of any specialized closed-set System One models. 

\section{Related Work}
\label{sec:related}

A general-purpose encoder maps an instruction, a context (text, image, video, etc.), and a set of candidates to a decision. Classification, retrieval, routing, reranking, and judgment all fit this one interface. They differ only in where the
candidate set comes from: a global corpus, a fixed label space, or a set defined by each
example, as in Jev-style tasks where every row carries its own options and criteria. Early
universal encoders focused on text retrieval and semantic similarity rather than instruction following~\citep{reimers2019sbert,
karpukhin2020dpr,wang2022e5,li2023gte,xiao2023bge}. Later work added instructions
to the input~\citep{su2023instructor,wang2024e5mistral,lee2024nvembed,muennighoff2024gritlm,
behnamghader2024llm2vec} and extended the method to multimodal inputs with vision-language
backbones~\citep{jiang2024e5v,jiang2024vlm2vec,zhang2024gme,lin2025mm,gu2026unime,chen2026douyin,zhang2024gme}, and augmenting encoder input with reasoning~\citep{cui2026think,cui2025reason,lan2026ume,cheng2026tte,jiang2026embed}. Benchmarks such as
MTEB~\citep{muennighoff2023mteb} and MMEB~\citep{jiang2024vlm2vec} already cast classification,
VQA, and grounding as scoring over candidates. Broadly, a model built on a language-model
backbone can score candidates in three ways. They differ in how much the prompt and the
candidates interact, and therefore in cost, scalability, and expressiveness:

\textbf{(1) Bi-encoders.}
A bi-encoder embeds the request (instruction, context, etc.)~and each candidate separately and scores
them by cosine similarity. It is trained with a  contrastive objective that pulls the
request representation toward its gold candidate and pushes it away from in-batch or hard
negatives~\citep{radford2021clip,jia2021align,zhai2023siglip,oord2018infonce}. This is the
dominant design for universal embedders~\citep{wang2022e5,su2023instructor,wang2024e5mistral,
lee2024nvembed,jiang2024vlm2vec,zhang2024gme}. Its main advantage is efficiency: since candidate
embeddings do not depend on the request,  they can be precomputed and indexed. Scoring $K$
candidates then costs a single forward pass for the request plus $K$ dot products, and this scales to
candidates of massive size. However, the tradeoff lies in expressiveness: since all interaction happens through a single
dot product,  the request embedding must anticipate every candidate it might be compared
against. This is a known weakness on fine-grained or compositional
distinctions~\citep{yuksekgonul2023aro,thrush2022winoground}.  Late-interaction models such as
ColBERT~\citep{khattab2020colbert} relax the single-vector bottleneck but still encode the two
sides independently. To address this limitation for closed-set decision making tasks, MetaEncoder places all option criteria into the request prompt to inject option-aware attention into the request representation. 

\textbf{(2) Cross-encoders.}
A cross-encoder concatenates the prompt with one candidate at a time. A shared
transformer processes the pair jointly, and a scalar relevance score is read from a
binary classification head or from the logit of a \textit{yes/true}
token~\citep{nogueira2019monobert,nogueira2020monot5,ma2023rankllama,zhang2025qwen3embedding}.
Full token-level attention between prompt and candidate makes cross-encoders consistently more
accurate than bi-encoders. That is why they serve as the second-stage reranker in
retrieve-then-rerank pipelines~\citep{nogueira2019monobert,xiao2023bge,cui2025reason}. LLM-as-a-judge
pointwise scoring follows the same pattern~\citep{zheng2023judge}. However, there are two costs following from the
design: First, no candidate-side representations can be cached, since each of the $K$ candidates needs to be concatenated with the prompt for scoring.
Second, each candidate is scored in isolation, so the model never compares candidates directly. 

\textbf{(3) Constrained decoding.}
The third approach puts the request and all candidates into a single
prompt. Each candidate is labeled with an identifier (A, B, \ldots, Z), and a generative model is asked to emit one identifier token. The prediction is the argmax of the next-token
distribution restricted to the valid identifiers. Since the vocabulary projection can be pruned to constrain outputs, the approach acts as a cheap encoder prefilling pass topped with a light linear decision layer. This is  the standard multiple-choice format
for evaluating LLMs~\citep{hendrycks2021mmlu,robinson2023mcsb}. It underlies zero-shot
classification through verbalizers~\citep{schick2021pet} and has been adopted for listwise
reranking, where the first generated identifier's logits act as the ranking
score~\citep{sun2023rankgpt,reddy2024first}. Because all candidates share one context, the
model can compare them jointly. Inference takes a single forward pass regardless of $K$, and
no new head or training objective is needed beyond the language-modeling loss. On the other hand, the
limitations come from treating candidates as tokens in a sequence. Accuracy is sensitive to
option order and identifier choice~\citep{pezeshkpour2023order,zheng2024pride}. Meanwhile the candidate
set is bounded by the size of the context window, so the approach cannot support large candidates. Besides, the output is a distribution over positions, but not
a reusable representation of the candidates. As a result, nothing can be indexed, cached across prompts,
or reused for retrieval and clustering.

\textbf{Summary.}
The three scoring methods trade off along one axis: how much the prompt and the candidates
interact. Bi-encoders allow shallower interactions, which buys indexing and scale. Cross-encoders let the prompt
interact with each candidate individually, which buys accuracy at  cost and no representation caching other than shared KV caches. Constrained
decoding lets all candidates interact with each other and with the prompt in one pass, but
it gives up open-ended candidate sets and reusable representations completely. 
\section{Scaling Dynamics of Candidate-Scoring Architectures}

\begin{figure}[t!]
    \centering
    \begin{subfigure}[b]{\linewidth}
        \centering
        \includegraphics[width=\linewidth]{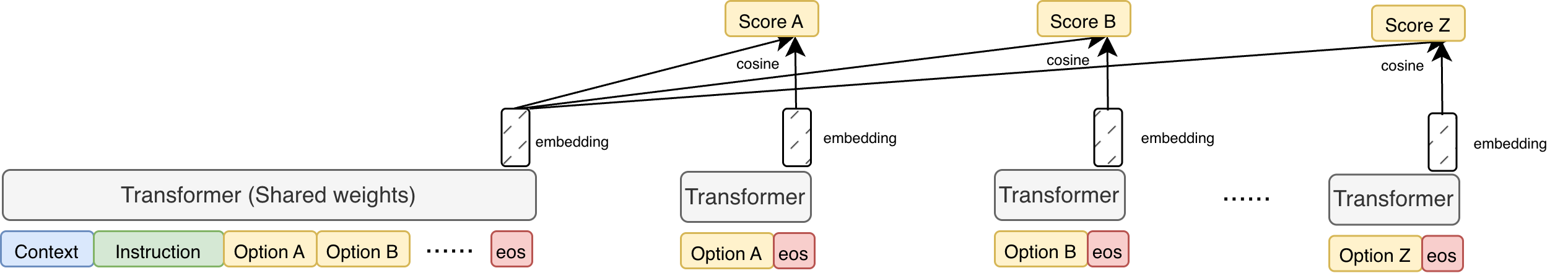}
        \caption{Bi-encoder. For closed-set options, all options are added to the query prompt.}
        \label{fig:sub1}
    \end{subfigure}

    \vspace{0.8em} 

    \begin{subfigure}[b]{0.55\linewidth}
        \centering
        \includegraphics[width=\linewidth]{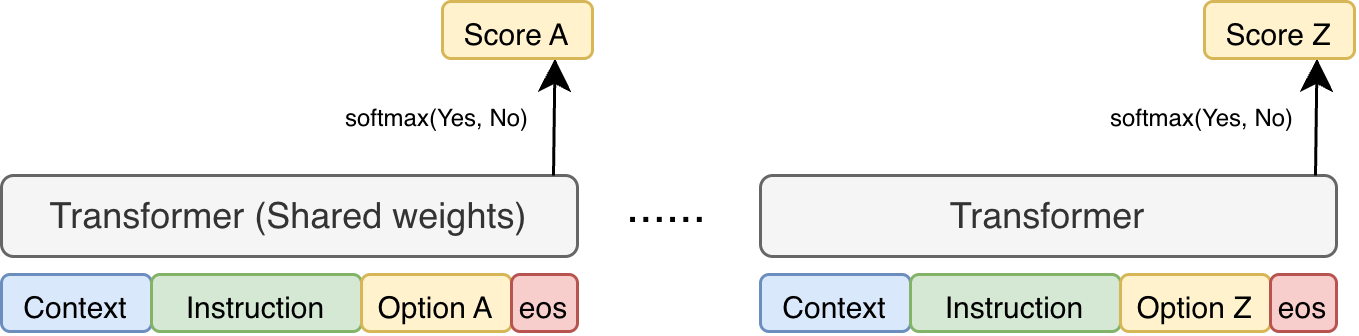}
        \caption{Cross-encoder.}
        \label{fig:sub2}
    \end{subfigure}\hfill
    \begin{subfigure}[b]{0.35\linewidth}
        \centering
        \includegraphics[width=\linewidth]{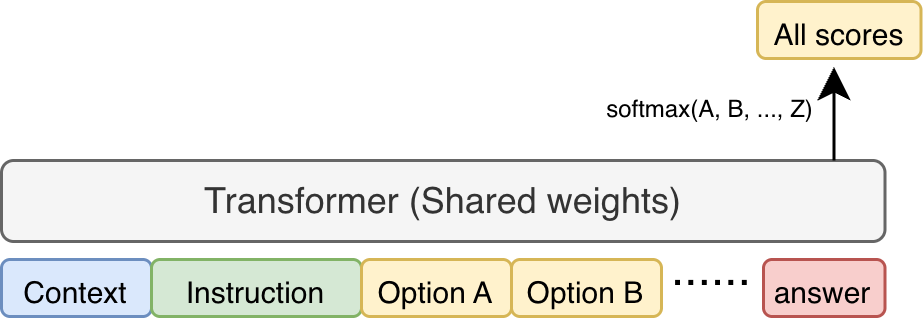}
        \caption{Constrained decoder.}
        \label{fig:sub3}
    \end{subfigure}

    \caption{A comparison between bi-encoder. cross encoder and constrained decoder for closed-set decision making.}
    \label{fig:main_figure}
\end{figure}

We study the scaling behavior of the three scoring mechanisms with respect to backbone decoder size and the number of training tokens. Models are trained on 10M request--candidate pairs in System One format and evaluated on 14K multiple-choice questions in a matching format. Figure~\ref{f2} illustrates the parameter and token scaling dynamics for each scorer, while Figure~\ref{fig:high_contrast_size_comparison} compares the three architectures across model scales.

\textbf{Scaling by model size and training tokens.} As shown in Figure~\ref{f2}, validation accuracy increases steadily with model size across all three scoring mechanisms. In contrast, scaling training tokens yields a sharp initial gain followed by more gradual improvements, as all scorers rapidly converge. A plausible explanation is that when adapting a pre-trained decoder for scoring, the underlying representational knowledge is already established; consequently, aligning the model to the scoring objective requires minimal additional training without compromising generalization.

\textbf{Comparison across scoring mechanisms.} As illustrated in Figure~\ref{fig:high_contrast_size_comparison}, the cross-encoder consistently outperforms the other two architectures across all model sizes. Between the bi-encoder and the constrained decoder, the bi-encoder achieves higher accuracy on smaller models (0.6B and 1.3B), but their performances converge as model size increases beyond 1.3B. This disparity likely stems from the fact that smaller decoders struggle to reliably produce constrained options when CoT reasoning is disabled, whereas larger decoders possess stronger inherent constrained-decoding capabilities. Notably, models with 4B or more parameters achieve zero-shot constrained decoding accuracy even before training; after fine-tuning, the performance gap between the bi-encoder and constrained decoder becomes marginal.

\begin{figure}[t!]
    \centering
    \includegraphics[width=1.0\textwidth]{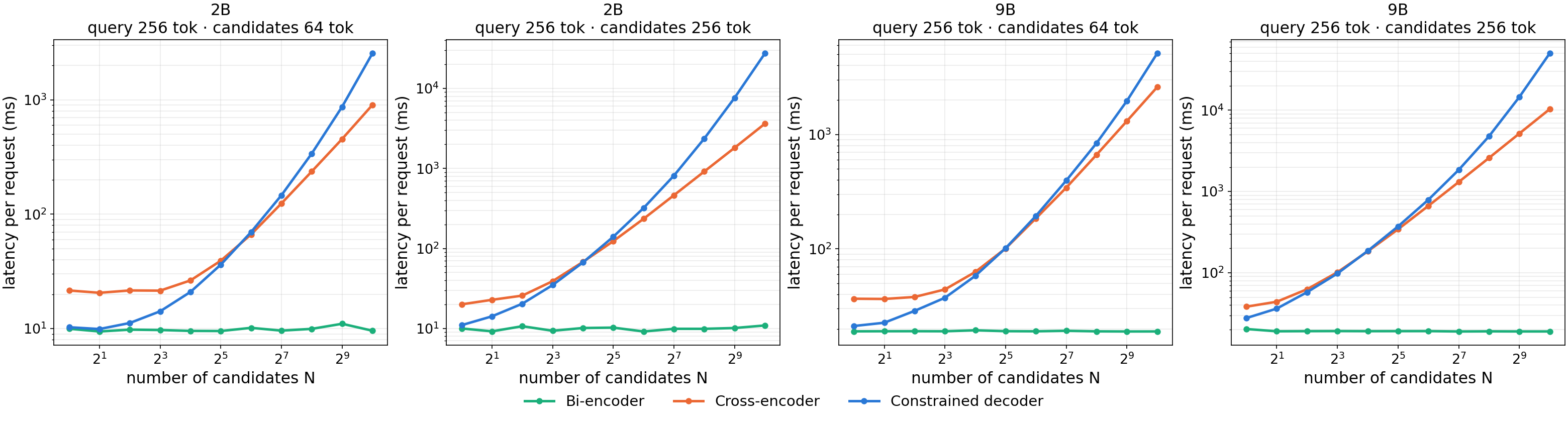}
    \caption{Scoring latency by number of candidates, measured with 1 x H100 and bf16. Two backbone sizes are measured: 2B and 9B. Two candidate lengths are measured: 64 tokens and 256 tokens.}
    \label{fig:latency}
\end{figure}

\textbf{Latency.} Figure~\ref{fig:latency} reports the per-request inference latency of each scoring architecture as candidate cardinality expands, evaluated on both 2B and 9B backbones. For the bi-encoder, candidate representations are pre-computed and indexed offline, requiring only a single forward pass over the query followed by vector dot products against all $N$ candidates. For the cross-encoder, we share the request KV cache across candidates and process all $N$ options in a single packed forward pass using a FlexAttention mask~\citep{dong2024flex}. Consequently, the bi-encoder maintains near-constant latency regardless of candidate pool size, whereas the latency of both the cross-encoder and constrained decoder scales quadratically with the number of candidates.

\begin{figure*}[t!]
\centering
\small
\begin{tikzpicture}
\begin{groupplot}[
    group style={
        group size=3 by 1,
        horizontal sep=1.4cm,
        y descriptions at=edge left,
    },
    width=5.8cm,
    height=4.8cm,
    clip=false,  
    grid=both,
    grid style={line width=0.15pt, draw=gray!20},
    major grid style={line width=0.3pt, draw=gray!40},
    xmin=0, xmax=7,
    ymin=0.2, ymax=1.0,
    ytick={0.2, 0.4, 0.6, 0.8, 1.0},
    xlabel={Training Step},
    xlabel style={font=\small},
    xtick={0,1,2,3,4,5,6,7},
    xticklabels={0, 1k, 2k, 3k, 4k, 5k, 6k, 7k},
    xticklabel style={font=\footnotesize},
    extra x ticks={0,1,2,3,4,5,6,7},
    extra x tick labels={0, 1.05, 2.10, 3.15, 4.19, 5.24, 6.29, 7.34},
    extra x tick style={
        tickpos=top,
        ticklabel pos=top,
        x tick label style={font=\scriptsize},
    },
    title style={font=\bfseries\normalsize, yshift=3.2em},
    legend style={
        font=\footnotesize,
        draw=none,
        legend columns=6,
        /tikz/every even column/.append style={column sep=0.8em},
    },
]

\nextgroupplot[
    title={(a) Bi-encoder},
    title style={font=\bfseries\footnotesize, yshift=-1.8em},
    ylabel={Accuracy},
    ylabel style={font=\small},
    legend to name=SizeLegend,
]
\node[anchor=south, font=\scriptsize] at (rel axis cs:0.5, 1.13) {Training Tokens (Billions)};

\addplot[thick, color=c700M, mark=*, mark size=1.3pt] coordinates {
    (0, 0.272) (1, 0.538) (2, 0.557) (3, 0.571) (4, 0.573) (5, 0.579) (6, 0.579) (7, 0.581)
}; \addlegendentry{0.6B}

\addplot[thick, color=c14B, mark=square*, mark size=1.3pt] coordinates {
    (0, 0.253) (1, 0.641) (2, 0.654) (3, 0.664) (4, 0.666) (5, 0.671) (6, 0.678) (7, 0.677)
}; \addlegendentry{1.3B}

\addplot[thick, color=c23B, mark=triangle*, mark size=1.5pt] coordinates {
    (0, 0.284) (1, 0.723) (2, 0.731) (3, 0.741) (4, 0.745) (5, 0.743) (6, 0.746) (7, 0.746)
}; \addlegendentry{2B}

\addplot[thick, color=c37B, mark=diamond*, mark size=1.6pt] coordinates {
    (0, 0.296) (1, 0.785) (2, 0.791) (3, 0.791) (4, 0.796) (5, 0.804) (6, 0.800) (7, 0.800)
}; \addlegendentry{4B}

\addplot[thick, color=c55B, mark=pentagon*, mark size=1.5pt] coordinates {
    (0, 0.300) (1, 0.818) (2, 0.823) (3, 0.819) (4, 0.825) (5, 0.822) (6, 0.827) (7, 0.828)
}; \addlegendentry{6B}

\addplot[thick, color=c9B, mark=star, mark size=1.6pt] coordinates {
    (0, 0.374) (1, 0.870) (2, 0.877) (3, 0.871) (4, 0.880) (5, 0.881) (6, 0.878) (7, 0.865)
}; \addlegendentry{9B}

\nextgroupplot[title={(b) Cross-encoder},
title style={font=\bfseries\footnotesize, yshift=-1.8em}]
\node[anchor=south, font=\scriptsize] at (rel axis cs:0.5, 1.13) {Training Tokens (Billions)};

\addplot[thick, color=c700M, mark=*, mark size=1.3pt] coordinates {
    (0, 0.319) (1, 0.583) (2, 0.593) (3, 0.603) (4, 0.607) (5, 0.609) (6, 0.610) (7, 0.613)
};
\addplot[thick, color=c14B, mark=square*, mark size=1.3pt] coordinates {
    (0, 0.294) (1, 0.673) (2, 0.684) (3, 0.688) (4, 0.690) (5, 0.694) (6, 0.702) (7, 0.701)
};
\addplot[thick, color=c23B, mark=triangle*, mark size=1.5pt] coordinates {
    (0, 0.370) (1, 0.772) (2, 0.781) (3, 0.786) (4, 0.783) (5, 0.788) (6, 0.791) (7, 0.793)
};
\addplot[thick, color=c37B, mark=diamond*, mark size=1.6pt] coordinates {
    (0, 0.272) (1, 0.807) (2, 0.814) (3, 0.814) (4, 0.820) (5, 0.820) (6, 0.829) (7, 0.830)
};
\addplot[thick, color=c55B, mark=pentagon*, mark size=1.5pt] coordinates {
    (0, 0.293) (1, 0.848) (2, 0.853) (3, 0.857) (4, 0.861) (5, 0.863) (6, 0.862) (7, 0.863)
};
\addplot[thick, color=c9B, mark=star, mark size=1.6pt] coordinates {
    (0, 0.371) (1, 0.908) (2, 0.904) (3, 0.909) (4, 0.909) (5, 0.909) (6, 0.910) (7, 0.908)
};

\nextgroupplot[title={(c) Constrained Decoder}, title style={font=\bfseries\footnotesize, yshift=-1.8em}]
\node[anchor=south, font=\scriptsize] at (rel axis cs:0.5, 1.13) {Training Tokens (Billions)};

\addplot[thick, color=c700M, mark=*, mark size=1.3pt] coordinates {
    (0, 0.232) (1, 0.505) (2, 0.523) (3, 0.538) (4, 0.544) (5, 0.548) (6, 0.559) (7, 0.558)
};
\addplot[thick, color=c14B, mark=square*, mark size=1.3pt] coordinates {
    (0, 0.230) (1, 0.633) (2, 0.643) (3, 0.647) (4, 0.652) (5, 0.658) (6, 0.663) (7, 0.665)
};
\addplot[thick, color=c23B, mark=triangle*, mark size=1.5pt] coordinates {
    (0, 0.228) (1, 0.715) (2, 0.734) (3, 0.741) (4, 0.753) (5, 0.756) (6, 0.763) (7, 0.762)
};
\addplot[thick, color=c37B, mark=diamond*, mark size=1.6pt] coordinates {
    (0, 0.687) (1, 0.796) (2, 0.809) (3, 0.811) (4, 0.817) (5, 0.816) (6, 0.818) (7, 0.816)
};
\addplot[thick, color=c55B, mark=pentagon*, mark size=1.5pt] coordinates {
    (0, 0.548) (1, 0.822) (2, 0.827) (3, 0.827) (4, 0.836) (5, 0.836) (6, 0.839) (7, 0.839)
};
\addplot[thick, color=c9B, mark=star, mark size=1.6pt] coordinates {
    (0, 0.808) (1, 0.859) (2, 0.861) (3, 0.859) (4, 0.866) (5, 0.857) (6, 0.856) (7, 0.854)
};

\end{groupplot}

\path (group c1r1.south west) -- (group c3r1.south east) 
    node[below=3.5em, midway] {\pgfplotslegendfromname{SizeLegend}};

\end{tikzpicture}
\caption{Scaling model size and number of training tokens for three scoring methods.}
\label{f2}
\end{figure*}
\definecolor{cBi}{HTML}{0072B2}      
\definecolor{cCross}{HTML}{D55E00}   
\definecolor{cDec}{HTML}{009E73}     

\begin{figure*}[t!]
\centering
\small
\begin{tikzpicture}
\begin{groupplot}[
    group style={
        group size=3 by 2,
        horizontal sep=0.9cm,
        vertical sep=1.2cm,
        y descriptions at=edge left,
    },
    width=5.8cm,
    height=4.6cm,
    clip=false,
    ymajorgrids=true,
    xmajorgrids=false,
    grid style={line width=0.25pt, draw=gray!25},
    xmin=0, xmax=7,
    xlabel style={font=\footnotesize},
    xtick={0,1,2,3,4,5,6,7},
    xticklabels={0, 1k, 2k, 3k, 4k, 5k, 6k, 7k},
    xticklabel style={font=\scriptsize},
    title style={font=\bfseries\footnotesize, yshift=0.3em},
    legend style={
        font=\footnotesize,
        draw=none,
        legend columns=3,
        /tikz/every even column/.append style={column sep=1.5em},
    },
    every axis plot/.append style={line width=1.1pt},
]


\nextgroupplot[
    title={0.6B}, 
    ylabel={Accuracy}, 
    ylabel style={font=\footnotesize},
    ymin=0.20, ymax=0.82,
    ytick={0.2, 0.4, 0.6, 0.8},
    yticklabel style={font=\scriptsize},
    extra x ticks={0,1,2,3,4,5,6,7},
    extra x tick labels={0, 1.05, 2.10, 3.15, 4.19, 5.24, 6.29, 7.34},
    extra x tick style={tickpos=top, ticklabel pos=top, x tick label style={font=\tiny}},
    legend to name=HighContrastLegend
]
\addplot[color=cBi, mark=*, mark size=1.5pt, mark options={fill=white, line width=0.8pt}] coordinates {
    (0, 0.272) (1, 0.538) (2, 0.557) (3, 0.571) (4, 0.573) (5, 0.579) (6, 0.579) (7, 0.581)
}; \addlegendentry{Bi-encoder}

\addplot[color=cCross, mark=square*, mark size=1.5pt, densely dashed, mark options={fill=white, line width=0.8pt}] coordinates {
    (0, 0.319) (1, 0.583) (2, 0.593) (3, 0.603) (4, 0.607) (5, 0.609) (6, 0.610) (7, 0.613)
}; \addlegendentry{Cross-encoder}

\addplot[color=cDec, mark=triangle*, mark size=1.8pt, dashdotdotted, mark options={fill=white, line width=0.8pt}] coordinates {
    (0, 0.232) (1, 0.505) (2, 0.523) (3, 0.538) (4, 0.544) (5, 0.548) (6, 0.559) (7, 0.558)
}; \addlegendentry{Constrained Decoder}

\nextgroupplot[
    title={1.3B},
    ymin=0.20, ymax=0.82,
    ytick={0.2, 0.4, 0.6, 0.8},
    extra x ticks={0,1,2,3,4,5,6,7},
    extra x tick labels={0, 1.05, 2.10, 3.15, 4.19, 5.24, 6.29, 7.34},
    extra x tick style={tickpos=top, ticklabel pos=top, x tick label style={font=\tiny}},
]
\node[anchor=south, font=\scriptsize\color{black!80}] at (rel axis cs:-0.05, 1.05) {Training Tokens (Billions)};

\addplot[color=cBi, mark=*, mark size=1.5pt, mark options={fill=white, line width=0.8pt}] coordinates {
    (0, 0.253) (1, 0.641) (2, 0.654) (3, 0.664) (4, 0.666) (5, 0.671) (6, 0.678) (7, 0.677)
};
\addplot[color=cCross, mark=square*, mark size=1.5pt, densely dashed, mark options={fill=white, line width=0.8pt}] coordinates {
    (0, 0.294) (1, 0.673) (2, 0.684) (3, 0.688) (4, 0.690) (5, 0.694) (6, 0.702) (7, 0.701)
};
\addplot[color=cDec, mark=triangle*, mark size=1.8pt, dashdotdotted, mark options={fill=white, line width=0.8pt}] coordinates {
    (0, 0.230) (1, 0.633) (2, 0.643) (3, 0.647) (4, 0.652) (5, 0.658) (6, 0.663) (7, 0.665)
};

\nextgroupplot[
    title={2B},
    ymin=0.20, ymax=0.82,
    ytick={0.2, 0.4, 0.6, 0.8},
    extra x ticks={0,1,2,3,4,5,6,7},
    extra x tick labels={0, 1.05, 2.10, 3.15, 4.19, 5.24, 6.29, 7.34},
    extra x tick style={tickpos=top, ticklabel pos=top, x tick label style={font=\tiny}},
]
\addplot[color=cBi, mark=*, mark size=1.5pt, mark options={fill=white, line width=0.8pt}] coordinates {
    (0, 0.284) (1, 0.723) (2, 0.731) (3, 0.741) (4, 0.745) (5, 0.743) (6, 0.746) (7, 0.746)
};
\addplot[color=cCross, mark=square*, mark size=1.5pt, densely dashed, mark options={fill=white, line width=0.8pt}] coordinates {
    (0, 0.370) (1, 0.772) (2, 0.781) (3, 0.786) (4, 0.783) (5, 0.788) (6, 0.791) (7, 0.793)
};
\addplot[color=cDec, mark=triangle*, mark size=1.8pt, dashdotdotted, mark options={fill=white, line width=0.8pt}] coordinates {
    (0, 0.228) (1, 0.715) (2, 0.734) (3, 0.741) (4, 0.753) (5, 0.756) (6, 0.763) (7, 0.762)
};


\nextgroupplot[
    title={4B}, 
    ylabel={Accuracy}, 
    ylabel style={font=\footnotesize},
    xlabel={Training Step},
    ymin=0.25, ymax=0.95,
    ytick={0.3, 0.5, 0.7, 0.9},
    yticklabel style={font=\scriptsize},
]
\addplot[color=cBi, mark=*, mark size=1.5pt, mark options={fill=white, line width=0.8pt}] coordinates {
    (0, 0.296) (1, 0.785) (2, 0.791) (3, 0.791) (4, 0.796) (5, 0.804) (6, 0.800) (7, 0.800)
};
\addplot[color=cCross, mark=square*, mark size=1.5pt, densely dashed, mark options={fill=white, line width=0.8pt}] coordinates {
    (0, 0.272) (1, 0.807) (2, 0.814) (3, 0.814) (4, 0.820) (5, 0.820) (6, 0.829) (7, 0.830)
};
\addplot[color=cDec, mark=triangle*, mark size=1.8pt, dashdotdotted, mark options={fill=white, line width=0.8pt}] coordinates {
    (0, 0.687) (1, 0.796) (2, 0.809) (3, 0.811) (4, 0.817) (5, 0.816) (6, 0.818) (7, 0.816)
};

\nextgroupplot[
    title={6B},
    xlabel={Training Step},
    ymin=0.25, ymax=0.95,
    ytick={0.3, 0.5, 0.7, 0.9},
]
\addplot[color=cBi, mark=*, mark size=1.5pt, mark options={fill=white, line width=0.8pt}] coordinates {
    (0, 0.300) (1, 0.818) (2, 0.823) (3, 0.819) (4, 0.825) (5, 0.822) (6, 0.827) (7, 0.828)
};
\addplot[color=cCross, mark=square*, mark size=1.5pt, densely dashed, mark options={fill=white, line width=0.8pt}] coordinates {
    (0, 0.293) (1, 0.848) (2, 0.853) (3, 0.857) (4, 0.861) (5, 0.863) (6, 0.862) (7, 0.863)
};
\addplot[color=cDec, mark=triangle*, mark size=1.8pt, dashdotdotted, mark options={fill=white, line width=0.8pt}] coordinates {
    (0, 0.548) (1, 0.822) (2, 0.827) (3, 0.827) (4, 0.836) (5, 0.836) (6, 0.839) (7, 0.839)
};

\nextgroupplot[
    title={9B},
    xlabel={Training Step},
    ymin=0.25, ymax=0.95,
    ytick={0.3, 0.5, 0.7, 0.9},
]
\addplot[color=cBi, mark=*, mark size=1.5pt, mark options={fill=white, line width=0.8pt}] coordinates {
    (0, 0.374) (1, 0.870) (2, 0.877) (3, 0.871) (4, 0.880) (5, 0.881) (6, 0.878) (7, 0.865)
};
\addplot[color=cCross, mark=square*, mark size=1.5pt, densely dashed, mark options={fill=white, line width=0.8pt}] coordinates {
    (0, 0.371) (1, 0.908) (2, 0.904) (3, 0.909) (4, 0.909) (5, 0.909) (6, 0.910) (7, 0.908)
};
\addplot[color=cDec, mark=triangle*, mark size=1.8pt, dashdotdotted, mark options={fill=white, line width=0.8pt}] coordinates {
    (0, 0.808) (1, 0.859) (2, 0.861) (3, 0.859) (4, 0.866) (5, 0.857) (6, 0.856) (7, 0.854)
};

\end{groupplot}

\path (group c1r2.south west) -- (group c3r2.south east) 
    node[below=2.0em, midway] {\pgfplotslegendfromname{HighContrastLegend}};

\end{tikzpicture}
\caption{Comparison of Bi-encoder, Cross-encoder, and Constrained Decoder architectures across 6 model sizes as a function of training step and token count.}
\label{fig:high_contrast_size_comparison}
\end{figure*}
\section{MetaEncoder}
\label{sec:model}

\subsection{Overview}
\label{sec:model:interface}

Guided by the above empirical study, we adopt a bi-encoder architecture to build \textbf{MetaEncoder}. Bi-encoders offer a compelling practical balance: candidate representations can be pre-computed and cached offline, allowing the model to scale efficiently to massive candidate pools, without substantially compromising closed-set decision accuracy. MetaEncoder scores a set of candidate options $\mathcal{C}=\{c_1,\dots,c_K\}$ against an incoming request $q$, where both requests and candidates can be multimodal. This architecture bridges closed-set decision-making and open-set retrieval within a single model, controlled entirely via prompt design. For closed-set tasks, the request prompt incorporates the task context, user instruction, and explicit candidate criteria, providing the request representation with option-aware context. For open-set retrieval across large corpora where enumerating all candidates in-context is intractable, the request prompt includes only the context and instruction. As a result, MetaEncoder produces both \textbf{candidate embeddings and scores against a request}.

\subsection{Architecture}
\label{sec:model:arch}

MetaEncoder is initialized from the pre-trained \texttt{Muse-Glimmer} decoder.
The same backbone $f_\theta$ encodes requests and candidates. Images and video are processed by
the backbone's native vision tower.

\textbf{Last-token representation.}
We keep the backbone's causal attention and take the final-layer hidden state of the
last token as the embedding of a request or a candidate, normalized to unit length:
\begin{equation}
  \mathbf{q} = \frac{h_{\text{last}}(q)}{\lVert h_{\text{last}}(q)\rVert_2}, \qquad
  \mathbf{c}_k = \frac{h_{\text{last}}(c_k)}{\lVert h_{\text{last}}(c_k)\rVert_2}.
\end{equation}
Under causal attention, the last token  attends to every token in the
input. It therefore sees the full context for the request, including state, instruction and optionally candidate criteria (for closed-set only), and can therefore summarize the request.

\textbf{Scoring.}
Each request–candidate pair is scored via cosine similarity, $s(q,c)=\mathbf{q}^{\top}\mathbf{c}$. Applying a softmax with temperature parameter $\tau$ over $\{s(q,c_k)/\tau\}_{k=1}^{K}$ produces a normalized probability distribution over the candidate pool. This mechanism scales efficiently from small closed sets to large corpora containing millions of candidates.

\subsection{Unidirectional Contrastive Learning}
\label{sec:model:loss}

We train MetaEncoder to align each request with its correct candidate. Each alignment is a pair
$(q_i, c_i^{+})$. For a batch of $B$ requests, the loss is an
InfoNCE objective~\citep{oord2018infonce} applied in the request-to-candidate direction only:
\begin{equation}
  \mathcal{L} = -\frac{1}{B}\sum_{i=1}^{B}
  \log \frac{\exp\!\big(s(q_i,c_i^{+})/\tau\big)}
            {\sum_{c\in\mathcal{N}(q_i)\cup\{c_i^{+}\}} \exp\!\big(s(q_i,c)/\tau\big)},
  \label{eq:loss}
\end{equation}
The negative set $\mathcal{N}(q_i)$ depends on the training task type:
\begin{itemize}
  \item \textbf{Open-set (retrieval):} $\mathcal{N}(q_i)$ consists of candidates from other requests gathered across the global batch, along with mined hard negatives where available. Any in-batch candidate identical to the positive target $c_i^{+}$ is masked out to avoid false negatives.
  \item \textbf{Closed-set (decision-making):} $\mathcal{N}(q_i)$ comprises the instance-specific candidate set excluding the gold target $c_i^{+}$. The softmax normalization is computed strictly over the options provided by the task, directly aligning the training objective with evaluation.
\end{itemize}

\subsection{Packed Contrastive Training}
\label{sec:packed}

Contrastive learning presents distinct scalability challenges across open-set and closed-set settings. In open-set retrieval, training demands large pools of in-batch negatives; however, scaling batch sizes for large vision-language models over multimodal inputs quickly saturates GPU activation memory. Prior solutions such as GradCache~\citep{jiang2024vlm2vec} decouple embedding generation from backward autograd graphs via a two-pass forward routine, but at the cost of substantial computational overhead. Conversely, closed-set scenarios involve a variable number of negative candidates per instance. Under conventional batching, naive padding suffers from compounding two-dimensional waste, as it must pad simultaneously to the maximum candidate count across instances and to the longest candidate sequence length.

To address both challenges with a hardware-efficient framework, we introduce Packed Contrastive Training. Rather than padding batches to the peak sequence length, we pack complete training instances into fixed 16,384-token sequences using first-fit bin packing. Each instance comprises a query, its positive candidate, and up to $k$ negative candidates. Instances are never partitioned across sequences, and those exceeding the context budget are discarded. By leveraging block-diagonal causal attention via FlexAttention~\citep{dong2024flex}, representations for each query and candidate are computed strictly over its own tokens. The Packed Contrastive Training
 fixes per-device token throughput rather than sample counts, so that the activation memory remains strictly invariant to sequence length variance. This eliminates padding overhead and decouples negative pool capacity from sequence length, enabling stable full fine-tuning of a 30B-parameter multimodal encoder with large, dynamic negative pools without gradient caching.

\section{Experiment}
\label{sec:eval}

\subsection{Benchmarks}
\label{sec:eval:benchmarks}

We evaluate MetaEncoder on 12 benchmarks, covering both closed-set candidate selection and open-set retrieval. The benchmark details are shown in Table~\ref{tab:benchmarks}.

\begin{table}[t]
\centering
\footnotesize
\begin{tabular}{llllrl}
\toprule
\textbf{Capability} & \textbf{Benchmark} & \textbf{Modality} & \textbf{Candidates} & \textbf{Size} & \textbf{Metric} \\
\midrule
\multirow{1}{*}{Decision making}
  & JEVBench  & Text       & Closed & 231   & Accuracy \\
\midrule
\multirow{6}{*}{\shortstack[l]{Multimodal\\understanding}}
  & MMLU                       & Text       & Closed & 14,033 & Hit@1\\
  & MMLU-Pro                       & Text       & Closed & 12,032 & Hit@1 \\
  & MMMU                       & Multimodal & Closed   & 847  & Hit@1 \\
  & MMMU-Pro                       & Multimodal & Closed   & 1,730  & Hit@1 \\
  & Video-MMMU                 & Multimodal & Closed   & 879   & Hit@1 \\
  & NaturalBench               & Multimodal & Closed & 7,598   & Hit@1 \\
  & TempCompass                & Multimodal & Closed   & 5,536   & Hit@1 \\
\midrule
\multirow{4}{*}{Retrieval}
  & NanoBEIR                   & Text       & Open   & 13 subsets  & nDCG@10 \\
  & MMEB-V3 Image              & Multimodal & Open   & 37 subsets  & Hit@1 \\
  & MMEB-V3 Video              & Multimodal & Open   & 18 subsets  & Hit@1 \\
  & MMEB-V3 VisDoc             & Multimodal & Open   & 24 subsets  & nDCG@5 \\
\bottomrule
\end{tabular}
\caption{Evaluation benchmarks. }
\label{tab:benchmarks}
\end{table}

\subsection{Evaluation}
\label{sec:eval:protocol}

\textbf{Closed set.}
Each request is scored only against its own options, and the prediction is the highest-scoring
option. We report accuracy, which is Hit@1 over the request's option set. 

\textbf{Open set.}
Each request is scored against the full candidate pool of its task. 
For NanoBEIR we report
nDCG@10 with linear gain. For MMEB-V3 we follow the benchmark's protocol and report Hit@1 on
the image and video partitions and nDCG@5 on visual documents.

\textbf{Comparisons}
We trained two models: MetaEncoder v0 and v1. Both models start with the same \texttt{Muse-Glimmer} backbone. MetaEncoder v0 is trained with  LoRA fine-tuning and GradCache~\citep{jiang2024vlm2vec}. MetaEncoder v1 is trained with full fine-tuning enabled by Packed Contrastive Training. For comparison, we consider two SOTA decision models: OpenJev-27B~\citep{openjev_2026} which a text-only cross-encoder variant  based on Qwen3.5-27B~\citep{team2026qwen3}, and Jev-Omni~\citep{akhilaaa3_jev_omni_2026} which is a multimodal constrained-decoder variant based on Gemma4-12B~\citep{team2026gemma}. We also consider two SOTA embedding models, WeMM-Embedding~\citep{zhou2026wemm} which holds first place on the MMEB-V3 leaderboard~\citep{tigerlab2026mmebleaderboard} and Qwen3-VL-Embedding~\citep{zhang2025qwen3embedding}.

\subsection{Results}
Table~\ref{tab:benchmark_results} reports the performance of MetaEncoder v0 and v1 across all evaluation suites relative to baseline models. MetaEncoder v1 achieves the highest score in 11 of the 14 benchmarks, demonstrating that a unified bi-encoder architecture can effectively support both multimodal decision-making and retrieval tasks. Relative to v0, the performance gains in MetaEncoder v1 stem from full fine-tuning paired with packed contrastive learning to maximize training efficiency. Comparing MetaEncoder and Jev models, the later models are limited by their inability to scale to open-set retrieval over large candidate pools. Compared MetaEncoder and SOTA embedding models, MetaEncoder v1 achieves SOTA image and video retrieval results on MMEB-V3. Notably, conventional SOTA embedding models struggle with closed-set answer retrieval, which is a critical performance gap that MetaEncoder effectively closes.

\begin{table*}[t]
\centering
\scriptsize
\begin{tabular}{lcccccc}
\toprule
\textbf{Benchmark} & \textbf{MetaEncoder v1} & \textbf{MetaEncoder v0} & \textbf{OpenJev} & \textbf{Jev-Omni} & \textbf{WeMM-Embedding (9B)} & \textbf{Qwen3-VL-Embedding (8B)} \\
\midrule
\multicolumn{7}{l}{\textit{Closed-set}} \\
\midrule
JevBench (orig)   & \textbf{1.000} & 0.944 & 0.986 & 0.972 & 0.625 & 0.639 \\
JevBench (easy)   & \textbf{1.000} & 1.000 & 1.000 & 0.917 & 0.979 & 0.938 \\
JevBench (hard)   & 0.685 & 0.739 & \textbf{0.766} & 0.739 & 0.487 & 0.432 \\
MMLU              & 0.837 & 0.777 & \textbf{0.860} & 0.773 & 0.733 & 0.654 \\
MMLU-Pro          & \textbf{0.671} & 0.561 & 0.572 & 0.552 & 3.000 & 0.412 \\
MMMU              & \textbf{0.665} & 0.579 & --    & 0.580 & 0.566 & 0.506 \\
MMMU-Pro          & \textbf{0.513} & 0.444 & --    & 0.375 & 0.415 & 0.329 \\
VideoMMMU         & \textbf{0.646} & 0.590 & --    & 0.571 & 0.485 & 0.419 \\
NaturalBench      & \textbf{0.832} & 0.813 & --    & 0.805 & 0.802 & 0.710 \\
TempCompass       & \textbf{0.778} & 0.735 & --    & 0.722 & 0.753 & 0.717 \\
\midrule
\multicolumn{7}{l}{\textit{Open-set}} \\
\midrule
NanoBEIR          & \textbf{0.701} & 0.663 & --    & --    & 0.618 & 0.606 \\
MMEB V3 (image)   & \textbf{0.816} & 0.790 & --    & --    & 0.810 & 0.780 \\
MMEB V3 (video)   & \textbf{0.743} & 0.649 & --    & --    & 0.743 & 0.672 \\
MMEB V3 (visdoc)  & 0.827 & 0.815 & --    & --    & \textbf{0.833} & 0.824 \\
\bottomrule
\end{tabular}
\caption{Benchmark Evaluation Results for MetaEncoder compared to SOTA System One decision makers and SOTA
retrieval models.}
\label{tab:benchmark_results}
\end{table*}

\subsection{Analysis}
\subsubsection{On the Importance of Adding Options in Request}
\label{sec:results:ablation-options}

Bi-encoders like MetaEncoder encode requests and candidates independently, computing scores via inner product instead of cross-attention. 
To handle multi-choice decision questions, MetaEncoder adds all options to the request prompt to inject option attention to the request representation, We study the importance of adding options to the prompt in Figure~\ref{option_list}. We find that \textbf{appending option lists to request yields substantial, uniform gains} across all MMLU, MMMU and Video-MMMU datasets.

\begin{figure}
\centering
\begin{tikzpicture}[
    x=8.5cm,
    y=0.95cm,
    font=\sffamily\small,
    barStyle/.style={rounded corners=1.5pt},
    barWith/.style={barStyle, fill=teal!65!black},
    barWithout/.style={barStyle, fill=orange!85!black},
    gapBar/.style={
        rounded corners=1.5pt,
        fill=red!8,
        draw=red!45,
        densely dashed,
        line width=0.45pt
    },
    gapText/.style={
        font=\scriptsize\bfseries\sffamily,
        text=red!75!black,
        fill=red!8,
        inner xsep=1.5pt,
        inner ysep=0.8pt,
        rounded corners=1.5pt
    }
]

  \foreach \x in {0.0, 0.2, 0.4, 0.6, 0.8, 1.0} {
    \draw[gray!20, thin] (\x, 0.35) -- (\x, 8.55);
    \draw[gray!60, thin] (\x, 0.35) -- (\x, 0.22)
      node[below, font=\footnotesize\sffamily, text=gray!80!black] {\x};
  }

  \draw[gray!60, -latex] (0, 0.35) -- (1.03, 0.35);
  \node[font=\footnotesize\bfseries\sffamily, text=gray!85!black] at (0.5, -0.45) {Score};
  \draw[gray!40] (0, 0.35) -- (0, 8.55);

  \fill[barWith] (0.00, 9.15) rectangle +(0.035, 0.22);
  \node[anchor=west, font=\footnotesize\sffamily] at (0.045, 9.26) {With option list};

  \fill[barWithout] (0.32, 9.15) rectangle +(0.035, 0.22);
  \node[anchor=west, font=\footnotesize\sffamily] at (0.365, 9.26) {Without};

  \filldraw[gapBar] (0.56, 9.15) rectangle +(0.035, 0.22);
  \node[anchor=west, font=\footnotesize\sffamily] at (0.605, 9.26) {Gap ($\Delta$)};

  \foreach \name/\w/\wo/\diff/\y in {
    {MMLU}/0.830/0.633/-0.197/8,
    {MMLU-Pro}/0.663/0.475/-0.189/7,
    {MMMU}/0.651/0.531/-0.120/6,
    {MMMU-Pro standard}/0.520/0.388/-0.132/5,
    {MMMU-Pro vision}/0.453/0.423/-0.030/4,
    {Video-MMMU Perception}/0.679/0.527/-0.152/3,
    {Video-MMMU Comprehension}/0.561/0.463/-0.098/2,
    {Video-MMMU Adaptation}/0.492/0.354/-0.138/1
  } {
    \node[anchor=east, align=right, font=\small\sffamily] at (-0.02, \y) {\name};

    \fill[barWith] (0, \y+0.05) rectangle (\w, \y+0.27);

    \fill[barWithout] (0, \y-0.27) rectangle (\wo, \y-0.05);

    \filldraw[gapBar] (\wo, \y-0.27) rectangle (\w, \y-0.05);
    \node[gapText] at ({(\wo+\w)/2}, \y-0.16) {\diff};
  }

\end{tikzpicture}
\caption{Effect of appending candidate list into the request prompt. Removing the candidate list from request leads to consistent regressions across all datasets.}
\label{option_list}
\end{figure}



\subsubsection{Candidate Format in Target}
\label{sec:results:candidates}

Next, to investigate what information candidate embeddings must encode, we hold the request fixed and vary the candidate formats: bare label, description only and label+description, as shown in Table~\ref{effect_candidate}.
\begin{table}[h]
    \centering
    \small
    \renewcommand{\arraystretch}{1.3}
    \begin{tabular}{llccc}
        \toprule
        \textbf{Candidate rendering} & \textbf{Example} & \textbf{MMLU} & \textbf{MMMU} & \textbf{MMLU} \\
        & & \textbf{(MetaEncoder v0)} & \textbf{(v0)} & \textbf{(v1)} \\
        \midrule
        bare label & ``C'' & 0.6512 & 0.4983 & 0.4760 \\
        description only & ``red, oxidative, slow contracting.'' & 0.7456 & 0.5629 & 0.8320 \\
        label+description & ``C. red, oxidative, slow contracting.'' & \textbf{0.7767} & \textbf{0.5788} & \textbf{0.8360} \\
        \bottomrule
    \end{tabular}
    \caption{Effect of different candidate rendering format. Bare label does not work well.}
    \label{effect_candidate}
\end{table}

Across both benchmarks (MMLU and MMMU) and MetaEncoder variants (v0 and v1), our findings are consistent: incorporating candidate descriptions into candidate encodings is critical, while using both label and descriptions provides the most discriminative candidate embeddings. Although bare labels conceptually reduce the task to resolving a pointer in the request, the bi-encoder struggles with this abstraction. Instead, \textbf{the bi-encoder relies on the discriminative semantic signals provided by option definitions to distinguish between candidates}.

\subsubsection{Sensitivity to Candidate Order}
\label{sec:results:order}
The previous sections analyze the request and candidate representations. A natural question to ask is the robustness of candidate ordering in both the request and candidate sides, where the same candidate can be assigned with a different label letter. 
For the evaluation benchmarks above, we permute the option order for each candidate. We run 10 independent permutations so that the candidate orders are shuffled randomly in each permutation. 
Across the 10 permutations for all benchmarks, none of the standard deviation exceeds 0.003. We conclude that \textbf{the bi-encoder based model is robust to option orders}.

\subsubsection{Video understanding}
\label{sec:results:video}

For video understanding, MetaEncoder compresses all frames in the request into a single representation. We study the impact of increasing the number of frames on Video-MMMU.
As shown in Figure~\ref{fig:videommmu-frames}, overall accuracy improves sharply from  $8$ frames to $16$ frames ($+0.08$), before plateauing at $64$ frames. Nearly all of the performance gain is realized by $16$ frames. Because input tokens scale linearly with the frame budget, scaling to $64$ frames incurs an $8\times$ computational overhead over the $8$-frame baseline, but with no improvement over $16$ frames.

\textbf{The gain is primarily perceptual.}
The performance improvement is heavily concentrated in tasks where the visual evidence is directly visible on screen. Perception accuracy exhibits continuous, monotonic gains across all budgets. In contrast, higher-level reasoning saturates early: Comprehension and Adaptation both improve at 16 frames but slightly regress at 64 frames. This divergence indicates that while denser frame sampling provides richer visual cues for perception, it does not alleviate the reasoning bottleneck of encoder-based architectures on complex multimodal adaptation tasks.


\definecolor{vmPerc}{HTML}{2563EB}    
\definecolor{vmComp}{HTML}{D97706}    
\definecolor{vmAdap}{HTML}{059669}    
\definecolor{vmAll}{HTML}{0F172A}     
\definecolor{vmSubtext}{HTML}{64748B} 
\definecolor{vmGrid}{HTML}{E2E8F0}    
\definecolor{vmAxis}{HTML}{94A3B8}    

\begin{figure}[t]
\centering
\begin{tikzpicture}[
    x=1.15cm, 
    y=1.00cm, 
    font=\sffamily\small,
    numLabel/.style={
        font=\fontsize{6pt}{7pt}\selectfont\bfseries\sffamily,
        fill=white,
        inner sep=1pt,
        rounded corners=1pt
    }
]

  \foreach \v/\y in {0.45/0, 0.50/0.7, 0.55/1.4, 0.60/2.1, 0.65/2.8, 0.70/3.5, 0.75/4.2, 0.80/4.9, 0.85/5.6} {
    \draw[vmGrid, line width=0.45pt] (0,\y) -- (9,\y);
    \draw[vmAxis, line width=0.45pt] (0,\y) -- (-0.08,\y);
    \node[anchor=east, font=\scriptsize\sffamily, text=vmSubtext] at (-0.12,\y) {\v};
  }

  \foreach \x in {3, 6, 9} {
    \draw[vmGrid, line width=0.45pt] (\x, 0) -- (\x, 5.6);
  }

  \draw[vmGrid, line width=0.45pt] (0, 5.6) -- (9, 5.6) -- (9, 0);

  \draw[vmAxis, line width=0.7pt] (0, 5.6) -- (0, 0) -- (9, 0);

  \begin{scope}[shift={(0, 6.18)}]
    \draw[vmPerc, line width=1.1pt] (0.25, 0) -- (0.65, 0);
    \fill[vmPerc] (0.45, 0) circle (2.2pt);
    \node[anchor=west, font=\scriptsize\bfseries\sffamily, text=vmPerc] at (0.75, 0) {Perception};

    \draw[vmComp, line width=1.1pt] (2.55, 0) -- (2.95, 0);
    \fill[vmComp] (2.75, 0) ++(0,2.5pt) -- ++(2.3pt,-3.9pt) -- ++(-4.6pt,0) -- cycle;
    \node[anchor=west, font=\scriptsize\bfseries\sffamily, text=vmComp] at (3.05, 0) {Comprehension};

    \draw[vmAdap, line width=1.1pt] (5.45, 0) -- (5.85, 0);
    \fill[vmAdap] (5.65, 0) ++(0,2.5pt) -- ++(2.3pt,-2.5pt) -- ++(-2.3pt,-2.5pt) -- ++(-2.3pt,2.5pt) -- cycle;
    \node[anchor=west, font=\scriptsize\bfseries\sffamily, text=vmAdap] at (5.95, 0) {Adaptation};

    \draw[vmAll, line width=1.8pt] (7.70, 0) -- (8.10, 0);
    \fill[vmAll] (7.90, 0) circle (2.8pt);
    \fill[white] (7.90, 0) circle (1.2pt);
    \node[anchor=west, font=\scriptsize\bfseries\sffamily, text=vmAll] at (8.20, 0) {Overall};
  \end{scope}

  \foreach \val [count=\i, evaluate=\i as \x using {int((\i-1)*3)}] in {8, 16, 32, 64} {
    \draw[vmAxis, line width=0.5pt] (\x,0) -- (\x,-0.08);
    \node[anchor=north, font=\scriptsize\bfseries\sffamily, text=vmAll] at (\x,-0.10) {\val};
  }

  \node[font=\footnotesize\bfseries\sffamily, text=vmAll] at (4.5,-0.75) {Sampled Frames};
  \node[font=\footnotesize\bfseries\sffamily, text=vmAll, rotate=90] at (-0.85, 2.80) {Accuracy};

  \draw[vmPerc, line width=1.1pt]
    (0,3.2060) -- (3,4.7600) -- (6,4.9000) -- (9,4.9700);
  \draw[vmComp, line width=1.1pt]
    (0,1.5540) -- (3,2.6320) -- (6,2.5900) -- (9,2.4500);
  \draw[vmAdap, line width=1.1pt]
    (0,0.5880) -- (3,1.3580) -- (6,1.1900) -- (9,1.1620);
  \draw[vmAll, line width=1.8pt]
    (0,1.8200) -- (3,2.9540) -- (6,2.9400) -- (9,2.8980);

  \foreach \x/\y in {0/3.2060, 3/4.7600, 6/4.9000, 9/4.9700}
    \fill[vmPerc] (\x,\y) circle (2.2pt);

  \foreach \x/\y in {0/1.5540, 3/2.6320, 6/2.5900, 9/2.4500}
    \fill[vmComp] (\x,\y) ++(0,2.5pt) -- ++(2.3pt,-3.9pt) -- ++(-4.6pt,0) -- cycle;

  \foreach \x/\y in {0/0.5880, 3/1.3580, 6/1.1900, 9/1.1620}
    \fill[vmAdap] (\x,\y) ++(0,2.5pt) -- ++(2.3pt,-2.5pt) -- ++(-2.3pt,-2.5pt) -- ++(-2.3pt,2.5pt) -- cycle;

  \foreach \x/\y in {0/1.8200, 3/2.9540, 6/2.9400, 9/2.8980} {
    \fill[vmAll] (\x,\y) circle (2.8pt);
    \fill[white] (\x,\y) circle (1.2pt);
  }

  \node[numLabel, text=vmPerc, anchor=south west, xshift=2pt, yshift=2.5pt] at (0,3.2060) {0.679};
  \node[numLabel, text=vmPerc, anchor=south, yshift=3.0pt] at (3,4.7600) {0.790};
  \node[numLabel, text=vmPerc, anchor=south, yshift=3.0pt] at (6,4.9000) {0.800};
  \node[numLabel, text=vmPerc, anchor=south east, xshift=-1pt, yshift=2.5pt] at (9,4.9700) {0.805};

  \node[numLabel, text=vmAll, anchor=south west, xshift=2pt, yshift=3.0pt] at (0,1.8200) {0.580};
  \node[numLabel, text=vmAll, anchor=south, yshift=3.5pt] at (3,2.9540) {0.661};
  \node[numLabel, text=vmAll, anchor=south, yshift=3.5pt] at (6,2.9400) {0.660};
  \node[numLabel, text=vmAll, anchor=south east, xshift=-1pt, yshift=3.0pt] at (9,2.8980) {0.657};

  \node[numLabel, text=vmComp, anchor=north west, xshift=2pt, yshift=-3.0pt] at (0,1.5540) {0.561};
  \node[numLabel, text=vmComp, anchor=north, yshift=-3.5pt] at (3,2.6320) {0.638};
  \node[numLabel, text=vmComp, anchor=north, yshift=-3.5pt] at (6,2.5900) {0.635};
  \node[numLabel, text=vmComp, anchor=north east, xshift=-1pt, yshift=-3.0pt] at (9,2.4500) {0.625};

  \node[numLabel, text=vmAdap, anchor=north west, xshift=2pt, yshift=-2.5pt] at (0,0.5880) {0.492};
  \node[numLabel, text=vmAdap, anchor=north, yshift=-3.0pt] at (3,1.3580) {0.547};
  \node[numLabel, text=vmAdap, anchor=north, yshift=-3.0pt] at (6,1.1900) {0.535};
  \node[numLabel, text=vmAdap, anchor=north east, xshift=-1pt, yshift=-2.5pt] at (9,1.1620) {0.533};

\end{tikzpicture}
\caption{Video-MMMU accuracy as a function of sampled frames. Doubling from 8 to 16 frames produces the steepest gain, plateauing at 32 and 64 frames. \emph{Overall} denotes the mean accuracy across all three tracks.}
\label{fig:videommmu-frames}
\end{figure}

\subsubsection{Missing Reasoning Ability}\label{sec:results:reasoning}

One hypothesis is that the intrinsic limitation of encoder models stems from their inability to perform  reasoning. To investigate this, we first evaluate the \texttt{Muse-Glimmer} backbone as a constrained decoder to directly predict a single option label without Chain-of-Thought (CoT). As shown in Figure~\ref{fig:cot_impact_mmmu}, direct prediction achieves an MMMU accuracy of only $0.49$, whereas enabling CoT substantially increases performance to $0.72$. This disparity suggests that the primary performance bottleneck in direct scoring models is the absence of intermediate reasoning.

\begin{figure}[t]
\centering
\begin{tikzpicture}
\begin{axis}[
    ybar=3pt,
    bar width=14pt,
    width=0.88\textwidth,
    height=7.2cm,
    ylabel={MMMU Accuracy},
    ylabel style={font=\bfseries, yshift=2pt},
    symbolic x coords={Decoder, MetaEncoder},
    xtick=data,
    xticklabels={
        \textbf{Decoder}\\[-1pt]\footnotesize(\texttt{Muse-Glimmer-30B}),
        \textbf{MetaEncoder}(v0)
    },
    x tick label style={align=center, font=\small},
    ymin=0.40, ymax=0.80,
    ytick={0.40, 0.45, 0.50, 0.55, 0.60, 0.65, 0.70, 0.75, 0.80},
    yticklabel style={
        /pgf/number format/fixed, 
        /pgf/number format/precision=2,
        /pgf/number format/zerofill
    },
    ymajorgrids=true,
    grid style={line width=0.3pt, draw=gray!25},
    axis line style={draw=gray!70, line width=0.6pt},
    axis x line*=bottom,
    axis y line*=left,
    enlarge x limits=0.55,
    nodes near coords,
    nodes near coords style={
        font=\fontseries{b}\selectfont\tiny,
        /pgf/number format/fixed,
        /pgf/number format/precision=2,
        /pgf/number format/zerofill
    },
    legend style={
        at={(0.5, 1.13)},
        anchor=south,
        legend columns=-1,
        draw=none,
        fill=none,
        font=\small,
        /tikz/every even column/.append style={column sep=14pt}
    }
]

\addplot[fill=colorDirect!85, draw=colorDirect!100, line width=0.5pt] 
coordinates {
    (Decoder, 0.49)
    (MetaEncoder, 0.57)
};

\addplot[fill=colorTruncated!85, draw=colorTruncated!100, line width=0.5pt] 
coordinates {
    (MetaEncoder, 0.68)
};

\addplot[fill=colorFull!85, draw=colorFull!100, line width=0.5pt] 
coordinates {
    (Decoder, 0.72)
    (MetaEncoder, 0.71)
};

\legend{Direct (No CoT), + Truncated CoT, + Full CoT}

\end{axis}
\end{tikzpicture}
\caption{Effect of Chain-of-Thought (CoT) reasoning for MMMU. Enabling CoT yields substantial improvements for both generation-based (Muse-Glimmer) and retrieval-based (MetaEncoder) multiple-choice question answering.}
\label{fig:cot_impact_mmmu}
\end{figure}

To determine whether bi-encoder-based architectures are bounded by the same constraint, we conduct additional experiments providing the CoT reasoning traces generated by \texttt{Muse-Glimmer} as auxiliary inputs to MetaEncoder~v0. As shown in Figure~\ref{fig:cot_impact_mmmu}, prepending the reasoning trace to the request boosts MetaEncoder's MMMU accuracy from $0.57$ to $0.71$. To verify that this gain is not simply an artifact of label leakage within the CoT trace, we also evaluate a setting where the final answer label is truncated from the CoT trace. In this cleaner setup, MetaEncoder still achieves an accuracy of $0.68$. These results demonstrate that intermediate reasoning is the dominant bottleneck for scoring models regardless of architecture, and incorporating reasoning traces effectively bridges this gap.

\section{Conclusion}
\label{sec:conclusion}
In this work, we systematically evaluated the trade-offs among bi-encoders, cross-encoders, and constrained decoders for candidate-scoring in System One decision-making tasks. Our experiments demonstrate that while cross-encoders consistently achieve the highest accuracy across model scales, bi-encoders offer a uniquely compelling operational profile: offline candidate caching maintains near-constant inference latency as candidate pools grow, while candidate-conditioned prompting substantially narrows the closed-set accuracy gap with joint-attention models. Guided by these insights, we introduced \textbf{MetaEncoder}, a multimodal encoder adapted from a pre-trained decoder via unidirectional contrastive learning and packed contrastive training to maximize training effeciency. By adopting a shared natural-language interface, MetaEncoder unifies closed-set categorical decisions and large-scale open retrieval within a single encoder model.

Evaluations show that MetaEncoder consistently outperforms existing multimodal encoders on decision tasks while being SOTA on image and video retrieval. Notably, candidate pool cardinality is not the primary determinant of task difficulty for MetaEncoder. Instead, performance boundaries are dictated by whether candidate distinctions possess clean semantic boundaries or demand pre-retrieval reasoning. Consequently, while MetaEncoder excels at perception and visual judgment, it shows clear limits on tasks requiring complex compositional deduction.
Ultimately, our analysis reveals that candidate-scoring models face an inherent ceiling on complex reasoning tasks compared to chain-of-thought generation. Bridging this performance gap without forfeiting the low-latency advantages of non-autoregressive scoring remains an important challenge.

\clearpage
\newpage
\bibliographystyle{assets/plainnat}
\bibliography{paper}

\begin{thebibliography}{60}
\providecommand{\natexlab}[1]{#1}
\providecommand{\url}[1]{\texttt{#1}}
\expandafter\ifx\csname urlstyle\endcsname\relax
  \providecommand{\doi}[1]{doi: #1}\else
  \providecommand{\doi}{doi: \begingroup \urlstyle{rm}\Url}\fi

\bibitem[AI(2025)]{nanobeir2025}
Sionic AI.
\newblock Nano-beir: A multilingual information retrieval benchmark with
  quality-enhanced queries, 2025.
\newblock \url{https://huggingface.co/sionic-ai}.

\bibitem[BehnamGhader et~al.(2024)BehnamGhader, Adlakha, Mosbach, Bahdanau,
  Chapados, and Reddy]{behnamghader2024llm2vec}
Parishad BehnamGhader, Vaibhav Adlakha, Marius Mosbach, Dzmitry Bahdanau,
  Nicolas Chapados, and Siva Reddy.
\newblock {LLM2Vec}: Large language models are secretly powerful text encoders.
\newblock In \emph{Conference on Language Modeling (COLM)}, 2024.

\bibitem[Chen et~al.(2026)Chen, Li, Wang, Liu, Wang, Jiang, and
  Dou]{chen2026douyin}
Haonan Chen, Chu Li, Zhicheng Wang, Yuanwei Liu, Yuanjiang Wang, Shaohua Jiang,
  and Zhicheng Dou.
\newblock Douyin multimodal embedding model technical report.
\newblock \emph{arXiv preprint arXiv:2608.02148}, 2026.

\bibitem[Cheng et~al.(2026)Cheng, Wu, Zhang, Bao, Ahuja, Mishra, Yu, Gao, Xia,
  Guo, et~al.]{cheng2026tte}
Jianpeng Cheng, Xian Wu, Jiangfan Zhang, Wentao Bao, Chaitanya Ahuja,
  Shlok~Kumar Mishra, Hanchao Yu, Yang Gao, Fan Xia, Qi~Guo, et~al.
\newblock Tte-flash: Accelerating reasoning-based multimodal representations
  via think-then-embed tokens.
\newblock \emph{arXiv preprint arXiv:2605.16638}, 2026.

\bibitem[Cui et~al.(2025)Cui, Chen, Yu, Yuan, Wang, Mishra, Yu, Yang, Xiao,
  Lim, et~al.]{cui2025reason}
Xuanming Cui, Hong-You Chen, Hao Yu, Hao Yuan, Zihao Wang, Shlok~Kumar Mishra,
  Hanchao Yu, Yonghuan Yang, Jun Xiao, Ser-Nam Lim, et~al.
\newblock Reason to contrast: A cascaded multimodal retrieval framework.
\newblock \emph{arXiv preprint arXiv:2602.23369}, 2025.

\bibitem[Cui et~al.(2026)Cui, Cheng, Chen, Shukla, Awasthi, Pan, Ahuja, Mishra,
  Tian, Guo, et~al.]{cui2026think}
Xuanming Cui, Jianpeng Cheng, Hong-you Chen, Satya~Narayan Shukla, Abhijeet
  Awasthi, Xichen Pan, Chaitanya Ahuja, Shlok Mishra, Taipeng Tian, Qi~Guo,
  et~al.
\newblock Think then embed: Generative context improves multimodal embedding.
\newblock In \emph{International Conference on Learning Representations},
  volume 2026, pages 2690--2709, 2026.

\bibitem[Dong et~al.(2024)Dong, Feng, Guessous, Liang, and He]{dong2024flex}
Juechu Dong, Boyuan Feng, Driss Guessous, Yanbo Liang, and Horace He.
\newblock Flex attention: A programming model for generating optimized
  attention kernels.
\newblock \emph{arXiv preprint arXiv:2412.05496}, 2\penalty0 (3):\penalty0 4,
  2024.

\bibitem[Gangi~Reddy et~al.(2024)Gangi~Reddy, Doo, Xu, Sultan, Swain, Sil, and
  Ji]{reddy2024first}
Revanth Gangi~Reddy, JaeHyeok Doo, Yifei Xu, Md~Arafat Sultan, Deevya Swain,
  Avirup Sil, and Heng Ji.
\newblock {FIRST}: Faster improved listwise reranking with single token
  decoding.
\newblock In \emph{Proceedings of the 2024 Conference on Empirical Methods in
  Natural Language Processing}, pages 8642--8652, 2024.

\bibitem[Gemma-Team et~al.(2026)Gemma-Team, Abd, Aggarwal, Algayres, Andreev,
  Bachem, Ballantyne, Brick, C{\u{a}}rbune, Casbon, et~al.]{team2026gemma}
Gemma-Team, Sherif~El Abd, Vaibhav Aggarwal, Robin Algayres, Alek Andreev,
  Olivier Bachem, Ian Ballantyne, Cormac Brick, Victor C{\u{a}}rbune, Michelle
  Casbon, et~al.
\newblock Gemma 4 technical report.
\newblock \emph{arXiv preprint arXiv:2607.02770}, 2026.

\bibitem[Gu et~al.(2026)Gu, Yang, Zhang, An, Feng, Zhang, Cai, Deng, and
  Bing]{gu2026unime}
Tiancheng Gu, Kaicheng Yang, Kaichen Zhang, Xiang An, Ziyong Feng, Yueyi Zhang,
  Weidong Cai, Jiankang Deng, and Lidong Bing.
\newblock Unime-v2: Mllm-as-a-judge for universal multimodal embedding
  learning.
\newblock In \emph{Proceedings of the AAAI Conference on Artificial
  Intelligence}, volume~40, pages 21378--21386, 2026.

\bibitem[Hendrycks et~al.(2021)Hendrycks, Burns, Basart, Zou, Mazeika, Song,
  and Steinhardt]{hendrycks2021mmlu}
Dan Hendrycks, Collin Burns, Steven Basart, Andy Zou, Mantas Mazeika, Dawn
  Song, and Jacob Steinhardt.
\newblock Measuring massive multitask language understanding.
\newblock In \emph{International Conference on Learning Representations
  (ICLR)}, 2021.

\bibitem[Hu et~al.(2025)Hu, Wu, Pu, Xiao, Zhang, Yue, Li, and
  Liu]{hu2025videommmu}
Kairui Hu, Penghao Wu, Fanyi Pu, Wang Xiao, Yuanhan Zhang, Xiang Yue, Bo~Li,
  and Ziwei Liu.
\newblock {Video-MMMU}: Evaluating knowledge acquisition from multi-discipline
  professional videos.
\newblock \emph{arXiv preprint arXiv:2501.13826}, 2025.

\bibitem[Huang et~al.(2026)Huang, Lu, Su, Zhang, Jiang, Nie, Zou, Pfister,
  Chen, Zhang, et~al.]{huang2026mmeb}
Haohang Huang, Xuan Lu, Mingyi Su, Xuan Zhang, Ziyan Jiang, Ping Nie, Kai Zou,
  Tomas Pfister, Wenhu Chen, Wei Zhang, et~al.
\newblock Mmeb-v3: Measuring the performance gaps of omni-modality embedding
  models.
\newblock \emph{arXiv preprint arXiv:2604.23321}, 2026.

\bibitem[HuggingFace(2026{\natexlab{a}})]{akhilaaa3_jev_omni_2026}
HuggingFace.
\newblock Jev-omni.
\newblock \url{https://huggingface.co/akhilaaa3/Jev-Omni}, 2026{\natexlab{a}}.
\newblock Multimodal decision classifier built on Gemma 4 12B IT.

\bibitem[HuggingFace(2026{\natexlab{b}})]{openjev_2026}
HuggingFace.
\newblock Openjev.
\newblock \url{https://huggingface.co/openjev/openjev}, 2026{\natexlab{b}}.
\newblock Open-weights decision model for typed text, web, and screenshot
  decisions.

\bibitem[Jia et~al.(2021)Jia, Yang, Xia, Chen, Parekh, Pham, Le, Sung, Li, and
  Duerig]{jia2021align}
Chao Jia, Yinfei Yang, Ye~Xia, Yi-Ting Chen, Zarana Parekh, Hieu Pham, Quoc~V.
  Le, Yunhsuan Sung, Zhen Li, and Tom Duerig.
\newblock Scaling up visual and vision-language representation learning with
  noisy text supervision.
\newblock In \emph{Proceedings of the 38th International Conference on Machine
  Learning (ICML)}, 2021.

\bibitem[Jiang et~al.(2026)Jiang, Wang, Zhu, Lu, Qin, Wang, Wan, and
  Tang]{jiang2026embed}
Haonan Jiang, Yuji Wang, Yongjie Zhu, Xin Lu, Wenyu Qin, Meng Wang, Pengfei
  Wan, and Yansong Tang.
\newblock Embed-rl: Reinforcement learning for reasoning-driven multimodal
  embeddings.
\newblock In \emph{European Conference on Computer Vision}, pages 39--56.
  Springer, 2026.

\bibitem[Jiang et~al.(2024)Jiang, Song, Zhang, Huang, Deng, Sun, Zhang, Wang,
  and Zhuang]{jiang2024e5v}
Ting Jiang, Minghui Song, Zihan Zhang, Haizhen Huang, Weiwei Deng, Feng Sun,
  Qi~Zhang, Deqing Wang, and Fuzhen Zhuang.
\newblock {E5-V}: Universal embeddings with multimodal large language models.
\newblock \emph{arXiv preprint arXiv:2407.12580}, 2024.

\bibitem[Jiang et~al.(2025)Jiang, Meng, Yang, Yavuz, Zhou, and
  Chen]{jiang2024vlm2vec}
Ziyan Jiang, Rui Meng, Xinyi Yang, Semih Yavuz, Yingbo Zhou, and Wenhu Chen.
\newblock {VLM2Vec}: Training vision-language models for massive multimodal
  embedding tasks.
\newblock In \emph{International Conference on Learning Representations
  (ICLR)}, 2025.

\bibitem[Karpukhin et~al.(2020)Karpukhin, O{\u{g}}uz, Min, Lewis, Wu, Edunov,
  Chen, and Yih]{karpukhin2020dpr}
Vladimir Karpukhin, Barlas O{\u{g}}uz, Sewon Min, Patrick Lewis, Ledell Wu,
  Sergey Edunov, Danqi Chen, and Wen-tau Yih.
\newblock Dense passage retrieval for open-domain question answering.
\newblock In \emph{Proceedings of the 2020 Conference on Empirical Methods in
  Natural Language Processing (EMNLP)}, pages 6769--6781, 2020.

\bibitem[Khattab and Zaharia(2020)]{khattab2020colbert}
Omar Khattab and Matei Zaharia.
\newblock {ColBERT}: Efficient and effective passage search via contextualized
  late interaction over {BERT}.
\newblock In \emph{Proceedings of the 43rd International ACM SIGIR Conference
  on Research and Development in Information Retrieval}, pages 39--48, 2020.

\bibitem[Lan et~al.(2026)Lan, Niu, Meng, Zhou, and Su]{lan2026ume}
Zhibin Lan, Liqiang Niu, Fandong Meng, Jie Zhou, and Jinsong Su.
\newblock Ume-r1: Exploring reasoning-driven generative multimodal embeddings.
\newblock In \emph{International Conference on Learning Representations},
  volume 2026, pages 115779--115806, 2026.

\bibitem[Lee et~al.(2025)Lee, Roy, Xu, Raiman, Shoeybi, Catanzaro, and
  Ping]{lee2024nvembed}
Chankyu Lee, Rajarshi Roy, Mengyao Xu, Jonathan Raiman, Mohammad Shoeybi, Bryan
  Catanzaro, and Wei Ping.
\newblock {NV-Embed}: Improved techniques for training {LLMs} as generalist
  embedding models.
\newblock In \emph{International Conference on Learning Representations
  (ICLR)}, 2025.

\bibitem[Li et~al.(2024)Li, Lin, Peng, Nyandwi, Jiang, Ma, Khanuja, Krishna,
  Neubig, and Ramanan]{li2024naturalbench}
Baiqi Li, Zhiqiu Lin, Wenxuan Peng, Jean de~Dieu Nyandwi, Daniel Jiang, Zixian
  Ma, Simran Khanuja, Ranjay Krishna, Graham Neubig, and Deva Ramanan.
\newblock {NaturalBench}: Evaluating vision-language models on natural
  adversarial samples.
\newblock In \emph{Advances in Neural Information Processing Systems
  (NeurIPS)}, 2024.

\bibitem[Li et~al.(2026)Li, Zhang, Long, Chen, Song, Bai, Yang, Xie, Yang, Liu,
  et~al.]{li2026qwen3}
Mingxin Li, Yanzhao Zhang, Dingkun Long, Keqin Chen, Sibo Song, Shuai Bai,
  Zhibo Yang, Pengjun Xie, An~Yang, Dayiheng Liu, et~al.
\newblock Qwen3-vl-embedding and qwen3-vl-reranker: A unified framework for
  state-of-the-art multimodal retrieval and ranking.
\newblock \emph{arXiv preprint arXiv:2601.04720}, 2026.

\bibitem[Li et~al.(2023)Li, Zhang, Zhang, Long, Xie, and Zhang]{li2023gte}
Zehan Li, Xin Zhang, Yanzhao Zhang, Dingkun Long, Pengjun Xie, and Meishan
  Zhang.
\newblock Towards general text embeddings with multi-stage contrastive
  learning.
\newblock \emph{arXiv preprint arXiv:2308.03281}, 2023.

\bibitem[Lin et~al.(2025)Lin, Lee, Shoeybi, Lin, Catanzaro, and
  Ping]{lin2025mm}
Sheng-Chieh Lin, Chankyu Lee, Mohammad Shoeybi, Jimmy Lin, Bryan Catanzaro, and
  Wei Ping.
\newblock Mm-embed: Universal multimodal retrieval with multimodal llms.
\newblock In \emph{International Conference on Learning Representations},
  volume 2025, pages 44215--44234, 2025.

\bibitem[Liu et~al.(2024)Liu, Li, Liu, Wang, Ren, Li, Chen, Sun, and
  Hou]{liu2024tempcompass}
Yuanxin Liu, Shicheng Li, Yi~Liu, Yuxiang Wang, Shuhuai Ren, Lei Li, Sishuo
  Chen, Xu~Sun, and Lu~Hou.
\newblock {TempCompass}: Do video {LLMs} really understand videos?
\newblock In \emph{Findings of the Association for Computational Linguistics:
  ACL 2024}, pages 8731--8772, 2024.

\bibitem[Ma et~al.(2024)Ma, Wang, Yang, Wei, and Lin]{ma2023rankllama}
Xueguang Ma, Liang Wang, Nan Yang, Furu Wei, and Jimmy Lin.
\newblock Fine-tuning {LLaMA} for multi-stage text retrieval.
\newblock In \emph{Proceedings of the 47th International ACM SIGIR Conference
  on Research and Development in Information Retrieval}, pages 2421--2425,
  2024.

\bibitem[{Meta Superintelligence
  Lab}(2026)]{MetaSuperintelligenceLab2026MuseGlimmer}
{Meta Superintelligence Lab}.
\newblock Muse glimmer-30b model card.
\newblock \url{https://huggingface.co/meta-models/Muse-Glimmer-30B}, august
  2026.
\newblock Hugging Face Model Repository.

\bibitem[Muennighoff et~al.(2023)Muennighoff, Tazi, Magne, and
  Reimers]{muennighoff2023mteb}
Niklas Muennighoff, Nouamane Tazi, Lo{\"i}c Magne, and Nils Reimers.
\newblock {MTEB}: Massive text embedding benchmark.
\newblock In \emph{Proceedings of the 17th Conference of the European Chapter
  of the Association for Computational Linguistics}, pages 2014--2037, 2023.

\bibitem[Muennighoff et~al.(2025)Muennighoff, Su, Wang, Yang, Wei, Yu, Singh,
  and Kiela]{muennighoff2024gritlm}
Niklas Muennighoff, Hongjin Su, Liang Wang, Nan Yang, Furu Wei, Tao Yu,
  Amanpreet Singh, and Douwe Kiela.
\newblock Generative representational instruction tuning.
\newblock In \emph{International Conference on Learning Representations
  (ICLR)}, 2025.

\bibitem[Nogueira and Cho(2019)]{nogueira2019monobert}
Rodrigo Nogueira and Kyunghyun Cho.
\newblock Passage re-ranking with {BERT}.
\newblock \emph{arXiv preprint arXiv:1901.04085}, 2019.

\bibitem[Nogueira et~al.(2020)Nogueira, Jiang, and Lin]{nogueira2020monot5}
Rodrigo Nogueira, Zhiying Jiang, and Jimmy Lin.
\newblock Document ranking with a pretrained sequence-to-sequence model.
\newblock In \emph{Findings of the Association for Computational Linguistics:
  EMNLP 2020}, pages 708--718, 2020.

\bibitem[Pezeshkpour and Hruschka(2024)]{pezeshkpour2023order}
Pouya Pezeshkpour and Estevam Hruschka.
\newblock Large language models sensitivity to the order of options in
  multiple-choice questions.
\newblock In \emph{Findings of the Association for Computational Linguistics:
  NAACL 2024}, pages 2006--2017, 2024.

\bibitem[Qwen-Team(2026)]{team2026qwen3}
Qwen-Team.
\newblock Qwen3. 5-omni technical report.
\newblock \emph{arXiv preprint arXiv:2604.15804}, 2026.

\bibitem[Radford et~al.(2021)Radford, Kim, Hallacy, Ramesh, Goh, Agarwal,
  Sastry, Askell, Mishkin, Clark, Krueger, and Sutskever]{radford2021clip}
Alec Radford, Jong~Wook Kim, Chris Hallacy, Aditya Ramesh, Gabriel Goh,
  Sandhini Agarwal, Girish Sastry, Amanda Askell, Pamela Mishkin, Jack Clark,
  Gretchen Krueger, and Ilya Sutskever.
\newblock Learning transferable visual models from natural language
  supervision.
\newblock In \emph{Proceedings of the 38th International Conference on Machine
  Learning (ICML)}, 2021.

\bibitem[Reimers and Gurevych(2019)]{reimers2019sbert}
Nils Reimers and Iryna Gurevych.
\newblock Sentence-{BERT}: Sentence embeddings using {S}iamese {BERT}-networks.
\newblock In \emph{Proceedings of the 2019 Conference on Empirical Methods in
  Natural Language Processing and the 9th International Joint Conference on
  Natural Language Processing (EMNLP-IJCNLP)}, pages 3982--3992, 2019.

\bibitem[Robinson et~al.(2023)Robinson, Rytting, and Wingate]{robinson2023mcsb}
Joshua Robinson, Christopher~Michael Rytting, and David Wingate.
\newblock Leveraging large language models for multiple choice question
  answering.
\newblock In \emph{International Conference on Learning Representations
  (ICLR)}, 2023.

\bibitem[Schick and Sch{\"u}tze(2021)]{schick2021pet}
Timo Schick and Hinrich Sch{\"u}tze.
\newblock Exploiting cloze-questions for few-shot text classification and
  natural language inference.
\newblock In \emph{Proceedings of the 16th Conference of the European Chapter
  of the Association for Computational Linguistics: Main Volume}, pages
  255--269, 2021.

\bibitem[Standhartinger and Contributors(2026)]{standhartinger2026jevbench}
Florian Standhartinger and Contributors.
\newblock Jevbench: Metamorphic coherence testing for typed probabilistic
  decision models.
\newblock \url{https://github.com/fstandhartinger/jevbench}, september 2026.
\newblock Benchmark Repository and Evaluation Suite.

\bibitem[Su et~al.(2023)Su, Shi, Kasai, Wang, Hu, Ostendorf, Yih, Smith,
  Zettlemoyer, and Yu]{su2023instructor}
Hongjin Su, Weijia Shi, Jungo Kasai, Yizhong Wang, Yushi Hu, Mari Ostendorf,
  Wen-tau Yih, Noah~A. Smith, Luke Zettlemoyer, and Tao Yu.
\newblock One embedder, any task: Instruction-finetuned text embeddings.
\newblock In \emph{Findings of the Association for Computational Linguistics:
  ACL 2023}, pages 1102--1121, 2023.

\bibitem[Sun et~al.(2023)Sun, Yan, Ma, Wang, Ren, Chen, Yin, and
  Ren]{sun2023rankgpt}
Weiwei Sun, Lingyong Yan, Xinyu Ma, Shuaiqiang Wang, Pengjie Ren, Zhumin Chen,
  Dawei Yin, and Zhaochun Ren.
\newblock Is {C}hat{GPT} good at search? investigating large language models as
  re-ranking agents.
\newblock In \emph{Proceedings of the 2023 Conference on Empirical Methods in
  Natural Language Processing}, pages 14918--14937, 2023.

\bibitem[Thrush et~al.(2022)Thrush, Jiang, Bartolo, Singh, Williams, Kiela, and
  Ross]{thrush2022winoground}
Tristan Thrush, Ryan Jiang, Max Bartolo, Amanpreet Singh, Adina Williams, Douwe
  Kiela, and Candace Ross.
\newblock Winoground: Probing vision and language models for visio-linguistic
  compositionality.
\newblock In \emph{Proceedings of the IEEE/CVF Conference on Computer Vision
  and Pattern Recognition (CVPR)}, pages 5228--5238, 2022.

\bibitem[{TIGER-Lab}(2026)]{tigerlab2026mmebleaderboard}
{TIGER-Lab}.
\newblock Massive multimodal embedding benchmark (mmeb) leaderboard.
\newblock \url{https://huggingface.co/spaces/TIGER-Lab/MMEB-Leaderboard}, 2026.
\newblock Accessed: 2026-10-07.

\bibitem[{TypeSafe AI}(2026)]{TypeSafeAI2026a}
{TypeSafe AI}.
\newblock Introducing system one models \& jev.
\newblock \url{https://typesafe.ai}, september 2026.
\newblock Technical Announcement and Early Access Release.

\bibitem[van~den Oord et~al.(2018)van~den Oord, Li, and
  Vinyals]{oord2018infonce}
Aaron van~den Oord, Yazhe Li, and Oriol Vinyals.
\newblock Representation learning with contrastive predictive coding.
\newblock \emph{arXiv preprint arXiv:1807.03748}, 2018.

\bibitem[Wang et~al.(2022)Wang, Yang, Huang, Jiao, Yang, Jiang, Majumder, and
  Wei]{wang2022e5}
Liang Wang, Nan Yang, Xiaolong Huang, Binxing Jiao, Linjun Yang, Daxin Jiang,
  Rangan Majumder, and Furu Wei.
\newblock Text embeddings by weakly-supervised contrastive pre-training.
\newblock \emph{arXiv preprint arXiv:2212.03533}, 2022.

\bibitem[Wang et~al.(2024{\natexlab{a}})Wang, Yang, Huang, Yang, Majumder, and
  Wei]{wang2024e5mistral}
Liang Wang, Nan Yang, Xiaolong Huang, Linjun Yang, Rangan Majumder, and Furu
  Wei.
\newblock Improving text embeddings with large language models.
\newblock In \emph{Proceedings of the 62nd Annual Meeting of the Association
  for Computational Linguistics (Volume 1: Long Papers)}, pages 11897--11916,
  2024{\natexlab{a}}.

\bibitem[Wang et~al.(2024{\natexlab{b}})Wang, Ma, Zhang, Ni, Chandra, Guo, Ren,
  Arulraj, He, Jiang, et~al.]{wang2024mmlu}
Yubo Wang, Xueguang Ma, Ge~Zhang, Yuansheng Ni, Abhranil Chandra, Shiguang Guo,
  Weiming Ren, Aaran Arulraj, Xuan He, Ziyan Jiang, et~al.
\newblock Mmlu-pro: A more robust and challenging multi-task language
  understanding benchmark.
\newblock \emph{Advances in Neural Information Processing Systems},
  37:\penalty0 95266--95290, 2024{\natexlab{b}}.

\bibitem[Xiao et~al.(2024)Xiao, Liu, Zhang, Muennighoff, Lian, and
  Nie]{xiao2023bge}
Shitao Xiao, Zheng Liu, Peitian Zhang, Niklas Muennighoff, Defu Lian, and
  Jian-Yun Nie.
\newblock C-pack: Packed resources for general chinese embeddings.
\newblock In \emph{Proceedings of the 47th International ACM SIGIR Conference
  on Research and Development in Information Retrieval}, pages 641--649, 2024.

\bibitem[Yue et~al.(2024)Yue, Ni, Zhang, Zheng, Liu, Zhang, Stevens, Jiang,
  Ren, Sun, Wei, Yu, Yuan, Sun, Yin, Zheng, Yang, Liu, Huang, Sun, Su, and
  Chen]{yue2024mmmu}
Xiang Yue, Yuansheng Ni, Kai Zhang, Tianyu Zheng, Ruoqi Liu, Ge~Zhang, Samuel
  Stevens, Dongfu Jiang, Weiming Ren, Yuxuan Sun, Cong Wei, Botao Yu, Ruibin
  Yuan, Renliang Sun, Ming Yin, Boyuan Zheng, Zhenzhu Yang, Yibo Liu, Wenhao
  Huang, Huan Sun, Yu~Su, and Wenhu Chen.
\newblock {MMMU}: A massive multi-discipline multimodal understanding and
  reasoning benchmark for expert {AGI}.
\newblock In \emph{Proceedings of the IEEE/CVF Conference on Computer Vision
  and Pattern Recognition (CVPR)}, pages 9556--9567, 2024.

\bibitem[Yue et~al.(2025)Yue, Zheng, Ni, Wang, Zhang, Tong, Sun, Yu, Zhang,
  Sun, et~al.]{yue2025mmmu}
Xiang Yue, Tianyu Zheng, Yuansheng Ni, Yubo Wang, Kai Zhang, Shengbang Tong,
  Yuxuan Sun, Botao Yu, Ge~Zhang, Huan Sun, et~al.
\newblock Mmmu-pro: A more robust multi-discipline multimodal understanding
  benchmark.
\newblock In \emph{Proceedings of the 63rd Annual Meeting of the Association
  for Computational Linguistics (Volume 1: Long Papers)}, pages 15134--15186,
  2025.

\bibitem[Yuksekgonul et~al.(2023)Yuksekgonul, Bianchi, Kalluri, Jurafsky, and
  Zou]{yuksekgonul2023aro}
Mert Yuksekgonul, Federico Bianchi, Pratyusha Kalluri, Dan Jurafsky, and James
  Zou.
\newblock When and why vision-language models behave like bags-of-words, and
  what to do about it?
\newblock In \emph{International Conference on Learning Representations
  (ICLR)}, 2023.

\bibitem[Zhai et~al.(2023)Zhai, Mustafa, Kolesnikov, and Beyer]{zhai2023siglip}
Xiaohua Zhai, Basil Mustafa, Alexander Kolesnikov, and Lucas Beyer.
\newblock Sigmoid loss for language image pre-training.
\newblock In \emph{Proceedings of the IEEE/CVF International Conference on
  Computer Vision (ICCV)}, pages 11941--11952, 2023.

\bibitem[Zhang et~al.(2025{\natexlab{a}})Zhang, Zhang, Xie, Li, Dai, Long, Xie,
  Zhang, Li, and Zhang]{zhang2024gme}
Xin Zhang, Yanzhao Zhang, Wen Xie, Mingxin Li, Ziqi Dai, Dingkun Long, Pengjun
  Xie, Meishan Zhang, Wenjie Li, and Min Zhang.
\newblock Bridging modalities: Improving universal multimodal retrieval by
  multimodal large language models.
\newblock In \emph{Proceedings of the IEEE/CVF Conference on Computer Vision
  and Pattern Recognition (CVPR)}, pages 9274--9285, 2025{\natexlab{a}}.

\bibitem[Zhang et~al.(2025{\natexlab{b}})Zhang, Li, Long, Zhang, Lin, Yang,
  Xie, Yang, Liu, Lin, Huang, and Zhou]{zhang2025qwen3embedding}
Yanzhao Zhang, Mingxin Li, Dingkun Long, Xin Zhang, Huan Lin, Baosong Yang,
  Pengjun Xie, An~Yang, Dayiheng Liu, Junyang Lin, Fei Huang, and Jingren Zhou.
\newblock {Qwen3} embedding: Advancing text embedding and reranking through
  foundation models.
\newblock \emph{arXiv preprint arXiv:2506.05176}, 2025{\natexlab{b}}.

\bibitem[Zheng et~al.(2024)Zheng, Zhou, Meng, Zhou, and Huang]{zheng2024pride}
Chujie Zheng, Hao Zhou, Fandong Meng, Jie Zhou, and Minlie Huang.
\newblock Large language models are not robust multiple choice selectors.
\newblock In \emph{International Conference on Learning Representations
  (ICLR)}, 2024.

\bibitem[Zheng et~al.(2023)Zheng, Chiang, Sheng, Zhuang, Wu, Zhuang, Lin, Li,
  Li, Xing, Zhang, Gonzalez, and Stoica]{zheng2023judge}
Lianmin Zheng, Wei-Lin Chiang, Ying Sheng, Siyuan Zhuang, Zhanghao Wu, Yonghao
  Zhuang, Zi~Lin, Zhuohan Li, Dacheng Li, Eric~P. Xing, Hao Zhang, Joseph~E.
  Gonzalez, and Ion Stoica.
\newblock Judging {LLM}-as-a-judge with {MT-Bench} and chatbot arena.
\newblock In \emph{Advances in Neural Information Processing Systems (NeurIPS),
  Datasets and Benchmarks Track}, pages 46595--46623, 2023.

\bibitem[Zhou et~al.(2026)Zhou, Mei, Li, Wang, Rao, and Lyu]{zhou2026wemm}
Junjie Zhou, Ke~Mei, Lei Li, Tianyi Wang, Fengyun Rao, and Jing Lyu.
\newblock Wemm-embedding: Wechat multi-modal embedding technical report.
\newblock \emph{arXiv preprint arXiv:2608.24053}, 2026.

\end{thebibliography}

\clearpage
\newpage
\beginappendix
\section{Prompts}
\label{app:prompts}

This appendix gives the exact prompt used for every evaluation suite, aligning with MetaEncoder's natural language interface.

\subsection{Common template}
\label{app:prompts:template}

All requests and candidates are assembled into a two-turn chat similar to a instruction-following decoder:

\begin{quote}\small
\textbf{system:} \texttt{Represent the user's input.}\\
\textbf{user:} \texttt{\{request\}} \quad (plus any image or video content parts)
\end{quote}

The system turn is a constant, always applied to both prompt and candidate encodings. The request goes to the user turn, which includes user instructions, context descriptions, criteria etc.
Media are attached in the MetaEncoder input interface. The chat template is applied with
\texttt{add\_generation\_prompt=False}, so the sequence ends at the user turn's
\texttt{<|eot|>}, which is what last-token pooling reads.
Candidates are encoded with the same template above, unless noted below, carry only their own
text.

\subsection{Decision making}
\label{app:prompts:decision}

\textbf{JEVBench and ImaJEV-Bench.}
Each row carries its own task definition and label ontology, rendered by a single shared
function:

\begin{quote}\small
\texttt{\{state\}}\\
\texttt{Task: \{instruction\}}\\
\texttt{Criteria:}\\
\texttt{- \{label\}: \{meaning of the label\}}\\
\texttt{Options: a, b, c}
\end{quote}

The criteria block is what makes these rows transferable: the label vocabulary is defined in the
prompt, so a row whose label set the model has never seen is still answerable. Boolean fields
are rendered as \texttt{yes: \{description\}} and \texttt{no: \{description\}} rather than
\texttt{True} and \texttt{False}, since a bare \texttt{"True"} is an opaque token that carries
no information about what the row is asking.

\textbf{S1MB (Additional Eval).}
The instruction is taken from each decision's \texttt{instructions\_json} field and the case
from \texttt{state\_json}, rendered as \texttt{key: value} lines. Candidates are the
natural-language \texttt{criteria[].description\_json} strings. The source's criterion
identifiers (\texttt{"B"}, \texttt{"option\_3"}, \texttt{"0"}) are used only to resolve which
candidate is correct and never reach the model. One row is emitted per (case, decision) pair,
because a single case may carry several decisions with different answer spaces.

An agentic control example:

\begin{quote}\small
\texttt{Choose one permitted tactical maneuver using the measured scene. The simulator retains
its collision reflex and flight controller.}\\
\\
\texttt{adapter: drone}\\
\texttt{observation: \{"control\_policy": \{"braking\_deceleration\_mps2": 3, \ldots\}\}}
\end{quote}

with candidates such as \texttt{"climb: Climb over an obstacle within verified clearance"} and
\texttt{"gap\_right: Use the observed safe gap on the right"}.

For ordinal (\texttt{score}) decisions each criterion additionally carries a numeric
\texttt{value}, which is retained in a sidecar so that predictions can receive graded credit:

\begin{quote}\small
\texttt{constraint: Do not infer additional numbers from associations}\\
\texttt{question: How much of the claim does the source support?}\\
\texttt{unit: Explicit numerical claims}\\
\\
\texttt{claim: Revenue rose 20\% and profit rose 20\%}\\
\texttt{source: Revenue rose 20\%. Profit figures were not disclosed.}
\end{quote}

with candidates \texttt{"Source mentions neither revenue nor profit"} (value 0),
\texttt{"Mentions revenue or profit but confirms neither claimed number"} (1),
\texttt{"Confirms exactly one claimed number"} (2) and
\texttt{"Confirms both claimed numbers"} (3).

\subsection{Multimodal understanding}
\label{app:prompts:multimodal}

\textbf{Video-MMMU and TempCompass.}
These three share one template, with the options rendered into the request:

\begin{quote}\small
\texttt{Given a video and a question, select the most accurate answer from the
provided candidates. Return only the exact text of your chosen answer. Question:\{question\}}\\
\texttt{Options:}\\
\texttt{(A) \{option\}}\\
\texttt{(B) \{option\}} \ldots
\end{quote}

Candidates are the full option sentence together with its letter marker, for example
\texttt{"(A) To get her to trust him"}. A verbatim MVBench request:

\begin{quote}\small
\texttt{Given a video and a question, select the most accurate answer from the
provided candidates. Return only the exact text of your chosen answer. Question:Why did Castle
dress like a fairy when he was speaking to Emily?}\\
\texttt{Options:}\\
\texttt{(A) To get her to trust him}\\
\texttt{(B) He secretly loved fairies} \ldots
\end{quote}

\textbf{TempCompass} Options arrive letter-prefixed in the source
(\texttt{"B. A man is dunking a basketball."}). The gold is resolved by matching that prefix,
after which the prefix is stripped, so the candidate is the sentence alone. Video inputs use
8 uniformly sampled frames drawn from 64 extracted per clip.

\textbf{Video-MMMU} \emph{Adaptation} track additionally references a still image. Frames are passed
through a separate video channel and the still as its own image content part, so that both
reach the model, and the \texttt{<image 1>} placeholder in the question text is reduced to the
word \texttt{image}.

\textbf{MMLU and MMMU.}
The base request is:

\begin{quote}\small
\texttt{Answer the multiple-choice question by selecting the correct option.}\\
\texttt{question: \{question\}}
\end{quote}

In the options-in-prompt variant, marked $*$ in the results tables, the alternatives are
additionally rendered into the request:

\begin{quote}\small
\texttt{Options:}\\
\texttt{A. \{option\}}\\
\texttt{B. \{option\}} \ldots
\end{quote}

Candidates are the full option text in both variants, never the letter. MMMU's
\texttt{<image N>} placeholders are reduced to the word \texttt{image}, since the referenced
images travel as content parts; rows whose candidates consist \emph{only} of an image reference
are excluded, because such a candidate carries no matchable text. A verbatim MMLU$^*$ request:

\begin{quote}\small
\texttt{Answer the multiple-choice question by selecting the correct option.}\\
\texttt{question: Type I muscle fibres have the following characteristics:}\\
\texttt{Options:}\\
\texttt{A. white, glycolytic, slow contracting.}\\
\texttt{B. red, oxidative, fast contracting.}\\
\texttt{C. red, oxidative, slow contracting.}\\
\texttt{D. white, oxidative, slow contracting.}
\end{quote}

\textbf{NaturalBench.}
The dataset's own yes/no question forms the request, with the image attached as a content part;
the candidates are \texttt{Yes} and \texttt{No}. Each group pairs two images with two questions
whose answers flip between the images, so answering from the question alone cannot succeed on
all four cells.

\subsection{Retrieval}
\label{app:prompts:retrieval}

\textbf{NanoBEIR.}
Queries carry the task instruction for their subset; passages are encoded without an
instruction. The instructions are, verbatim:

\begin{quote}\small
\begin{tabular}{@{}ll@{}}
\toprule
Subset & Query instruction \\
\midrule
NanoMSMARCO        & Given a web search query, retrieve relevant passages that answer the query \\
NanoNQ             & Given a question, retrieve passages that answer the question \\
NanoHotpotQA       & Given a multi-hop question, retrieve documents that can help answer the question \\
NanoFEVER          & Given a claim, retrieve documents that support or refute the claim \\
NanoNFCorpus       & Given a question, retrieve relevant documents that best answer the question \\
NanoFiQA2018       & Given a financial question, retrieve user replies that best answer the question \\
NanoQuoraRetrieval & Given a question, retrieve questions that are semantically equivalent \\
NanoArguAna        & Given a claim, find documents that refute the claim \\
NanoClimateFEVER   & Given a claim about climate change, retrieve documents that support or refute the claim \\
NanoDBPedia        & Given a query, retrieve relevant entity descriptions from DBPedia \\
NanoSCIDOCS        & Given a scientific paper title, retrieve paper abstracts that are cited by the given paper \\
NanoSciFact        & Given a scientific claim, retrieve documents that support or refute the claim \\
NanoTouche2020     & Given a question, retrieve detailed and persuasive arguments that answer the question \\
\bottomrule
\end{tabular}
\end{quote}

\textbf{MMEB-V3.}
For MMEB-V3 each task ships its own query
instruction in the benchmark metadata, and we use it directly in prompt. The request is assembled as
\texttt{qry\_instuction} followed by the row's \texttt{qry\_text}, with the image attached as a
content part; candidates are built from \texttt{tgt\_text} and \texttt{tgt\_img\_path}.
Instructions are per task, and their form follows the
task type: classification tasks ask for a representation of the image, retrieval tasks describe
what should be retrieved, and grounding tasks describe the region to select.

\begin{quote}\small
\begin{tabular}{@{}p{0.19\linewidth}p{0.75\linewidth}@{}}
\toprule
Task & \texttt{qry\_inst} \\
\midrule
ImageNet-1K & \texttt{Represent the given image for classification} \\
Country211  & \texttt{Identify the country depicted in the image} \\
DocVQA      & \texttt{Represent the given image with the following question:} \\
RefCOCO     & \texttt{Select the portion of the image that follows the language expressions.} \\
OVEN        & \texttt{Retrieve a Wikipedia image-description pair that provides evidence for the question of this image:} \\
FashionIQ   & \texttt{Find an image to match the fashion image and style note:} \\
MSCOCO\_t2i & \texttt{Find me an everyday image that matches the given caption:} \\
VisDial     & \texttt{Represent the given dialogue about an image, which is used for image retrieval:} \\
\bottomrule
\end{tabular}
\end{quote}

\clearpage
{\tiny
\begin{longtable}[l]{|l|l|c|}
\hline
\textbf{Task} & \textbf{Type} & \textbf{Acc} \\ \hline
\endhead

open\_jev\_\_browser\_control\_v1 & choice & 1 \\ \hline
laya\_\_enron\_spam & noul & 0.96 \\ \hline
miqa & choice & 0.96 \\ \hline
open\_jev\_\_amount\_extraction\_control\_v1 & noul & 0.96 \\ \hline
open\_jev\_\_mailroom\_control\_v1 & noul & 0.95 \\ \hline
winobias & choice & 0.94 \\ \hline
followir\_robust04 & noul & 0.935 \\ \hline
workflow\_controls\_\_customer\_service & noul & 0.93 \\ \hline
generalization\_contextual\_choice & choice & 0.93 \\ \hline
open\_jev\_\_silent\_failure\_control\_v1 & noul & 0.92 \\ \hline
hans & noul & 0.91 \\ \hline
go\_emotions & noul & 0.9085 \\ \hline
workflow\_controls\_\_invoice\_processing & noul & 0.9 \\ \hline
workflow\_controls\_\_security\_incidents & noul & 0.9 \\ \hline
hatecheck & noul & 0.89 \\ \hline
impli & noul & 0.89 \\ \hline
workflow\_controls\_\_agent\_trace\_observability & noul & 0.89 \\ \hline
laya\_\_ag\_news & choice & 0.88 \\ \hline
sdoh\_nli & noul & 0.88 \\ \hline
open\_jev\_\_painting\_geometry\_v1 & noul & 0.87 \\ \hline
generalization\_diverse\_choice & choice & 0.87 \\ \hline
defeasible\_nli & choice & 0.86 \\ \hline
open\_jev\_\_context\_retention\_control\_v1 & noul & 0.86 \\ \hline
creak & noul & 0.85 \\ \hline
dbpedia & choice & 0.85 \\ \hline
quartz & choice & 0.85 \\ \hline
snli & choice & 0.85 \\ \hline
synth\_rel\_\_NanoFEVER & score & 0.84 \\ \hline
synth\_rel\_\_NanoSciFact & score & 0.8387 \\ \hline
paws & noul & 0.83 \\ \hline
scitail & noul & 0.83 \\ \hline
arc & choice & 0.81 \\ \hline
babi\_nli & noul & 0.81 \\ \hline
open\_jev\_\_customer\_control\_v1 & score & 0.81 \\ \hline
open\_jev\_\_vizdoom\_basic\_v1 & score & 0.803 \\ \hline
synth\_rel\_\_NanoNQ & score & 0.8025 \\ \hline
synth\_rel\_\_nanocoir & score & 0.8 \\ \hline
clinc & choice & 0.7748 \\ \hline
ethos & noul & 0.77 \\ \hline
open\_jev\_\_phone\_extraction\_control\_v1 & choice & 0.77 \\ \hline
cladder & noul & 0.76 \\ \hline
followir\_news21 & noul & 0.76 \\ \hline
synth\_rel\_\_NanoHotpotQA & score & 0.755 \\ \hline
lonli & choice & 0.75 \\ \hline
gsm8k & choice & 0.74 \\ \hline
open\_jev\_\_ir\_control\_v1 & score & 0.74 \\ \hline
massive & choice & 0.735 \\ \hline
openbookqa & choice & 0.73 \\ \hline
synth\_rel\_\_NanoSCIDOCS & score & 0.7288 \\ \hline
synth\_rel\_\_NanoNFCorpus & score & 0.7275 \\ \hline
followir\_core17 & noul & 0.725 \\ \hline
few\_nerd & choice & 0.7235 \\ \hline
civil\_comments & noul & 0.7229 \\ \hline
hwu64 & choice & 0.72 \\ \hline
mtop & choice & 0.715 \\ \hline
canttalk & noul & 0.71 \\ \hline
generalization\_diverse\_noul & noul & 0.71 \\ \hline
aegis2 & noul & 0.7027 \\ \hline
synth\_rel\_\_NanoMSMARCO & score & 0.6925 \\ \hline
open\_jev\_\_sponsor\_segment\_control\_v1 & choice & 0.69 \\ \hline
synth\_rel\_\_NanoClimateFEVER & score & 0.69 \\ \hline
synth\_rel\_\_NanoArguAna & score & 0.6725 \\ \hline
open\_jev\_\_citation\_control\_v1 & choice & 0.67 \\ \hline
open\_jev\_\_email\_selection\_control\_v1 & noul & 0.67 \\ \hline
synth\_rel\_\_NanoQuoraRetrieval & score & 0.6687 \\ \hline
arct & choice & 0.66 \\ \hline
open\_jev\_\_entity\_alignment\_control\_v1 & noul & 0.66 \\ \hline
generalization\_contextual\_noul & noul & 0.66 \\ \hline
synth\_rel\_\_NanoDBPedia & score & 0.6462 \\ \hline
open\_jev\_\_reasoning\_control\_v1 & score & 0.64 \\ \hline
ruletaker & noul & 0.64 \\ \hline
scone & choice & 0.64 \\ \hline
ethics & noul & 0.63 \\ \hline
synth\_rel\_\_NanoFiQA2018 & score & 0.6275 \\ \hline
lexcomp & noul & 0.62 \\ \hline
plane & noul & 0.59 \\ \hline
aqua\_rat & choice & 0.58 \\ \hline
bbq & choice & 0.58 \\ \hline
scicite & choice & 0.58 \\ \hline
crows\_pairs & choice & 0.57 \\ \hline
robust\_lr & choice & 0.57 \\ \hline
contract\_nli & choice & 0.5676 \\ \hline
gretel\_pii & choice & 0.5666 \\ \hline
banking77 & choice & 0.56 \\ \hline
open\_jev\_\_snake\_v1 & noul & 0.56 \\ \hline
laya\_\_typed\_decisions & score & 0.556 \\ \hline
logical\_entailment & noul & 0.55 \\ \hline
generalization\_diverse\_score & score & 0.55 \\ \hline
temporal\_nli & choice & 0.55 \\ \hline
generalization\_contextual\_score & score & 0.53 \\ \hline
tracie & noul & 0.53 \\ \hline
hh\_rlhf & choice & 0.52 \\ \hline
winowhy & noul & 0.52 \\ \hline
sgd & choice & 0.51 \\ \hline
nlsat & noul & 0.5 \\ \hline
ud\_ewt & choice & 0.4944 \\ \hline
qasper & noul & 0.49 \\ \hline
synth\_rel\_\_NanoTouche2020 & score & 0.4809 \\ \hline
boardgameqa & choice & 0.48 \\ \hline
open\_jev\_\_drone\_control\_v1 & noul & 0.44 \\ \hline
fol\_nli & choice & 0.43 \\ \hline
wiqa & choice & 0.42 \\ \hline
corr2cause & choice & 0.41 \\ \hline
open\_jev\_\_tic\_tac\_toe\_v1 & choice & 0.38 \\ \hline
spartqa & choice & 0.37 \\ \hline
patent\_similarity & score & 0.36 \\ \hline
esci & choice & 0.33 \\ \hline
poem\_sentiment & choice & 0.33 \\ \hline
stepgame & choice & 0.28 \\ \hline
argument\_quality & noul & 0.27 \\ \hline

\caption{Full results on S1MB (110 tasks).} \label{tab:s1mb_full} \\
\end{longtable}
}

{\tiny
\begin{longtable}[l]{|l|c|}
\hline
\textbf{Task} & \textbf{Score} \\ \hline
\endhead

scene\_transition & 0.925 \\ \hline
unexpected\_action & 0.775 \\ \hline
moving\_attribute & 0.755 \\ \hline
object\_existence & 0.695 \\ \hline
character\_order & 0.675 \\ \hline
action\_prediction & 0.645 \\ \hline
object\_interaction & 0.615 \\ \hline
action\_sequence & 0.585 \\ \hline
state\_change & 0.565 \\ \hline
fine\_grained\_pose & 0.56 \\ \hline
action\_antonym & 0.55 \\ \hline
fine\_grained\_action & 0.52 \\ \hline
moving\_direction & 0.48 \\ \hline
episodic\_reasoning & 0.465 \\ \hline
action\_localization & 0.425 \\ \hline
counterfactual\_inference & 0.395 \\ \hline
object\_shuffle & 0.375 \\ \hline
action\_count & 0.37 \\ \hline
moving\_count & 0.35 \\ \hline
egocentric\_navigation & 0.35 \\ \hline
\caption{Full results on MVBench (20 tasks).} \label{tab:full_mvbench} \\
\end{longtable}
}

{\tiny
\begin{longtable}[l]{|l|c|c|}

\hline
\textbf{Subject} & \textbf{Score (options not in prompt)} & \textbf{Score (options in prompt)} \\ \hline
\endhead
security\_studies & 0.2939 & 0.702 \\ \hline
international\_law & 0.405 & 0.7769 \\ \hline
human\_sexuality & 0.4733 & 0.8168 \\ \hline
nutrition & 0.4379 & 0.7778 \\ \hline
moral\_disputes & 0.4335 & 0.7139 \\ \hline
jurisprudence & 0.4907 & 0.7685 \\ \hline
high\_school\_microeconomics & 0.5672 & 0.8361 \\ \hline
sociology & 0.597 & 0.8607 \\ \hline
high\_school\_physics & 0.4702 & 0.7285 \\ \hline
college\_biology & 0.6042 & 0.8542 \\ \hline
econometrics & 0.4035 & 0.6404 \\ \hline
high\_school\_chemistry & 0.4483 & 0.6798 \\ \hline
high\_school\_government\_and\_politics & 0.6788 & 0.9067 \\ \hline
clinical\_knowledge & 0.5434 & 0.766 \\ \hline
high\_school\_statistics & 0.5417 & 0.7639 \\ \hline
computer\_security & 0.56 & 0.78 \\ \hline
professional\_accounting & 0.4823 & 0.7021 \\ \hline
high\_school\_geography & 0.6717 & 0.8889 \\ \hline
astronomy & 0.5921 & 0.8092 \\ \hline
high\_school\_macroeconomics & 0.541 & 0.7564 \\ \hline
electrical\_engineering & 0.4621 & 0.6759 \\ \hline
us\_foreign\_policy & 0.58 & 0.79 \\ \hline
high\_school\_computer\_science & 0.63 & 0.84 \\ \hline
high\_school\_world\_history & 0.654 & 0.8565 \\ \hline
abstract\_algebra & 0.5 & 0.7 \\ \hline
philosophy & 0.4952 & 0.6945 \\ \hline
high\_school\_biology & 0.6677 & 0.8645 \\ \hline
high\_school\_us\_history & 0.7059 & 0.902 \\ \hline
high\_school\_european\_history & 0.6061 & 0.8 \\ \hline
college\_mathematics & 0.45 & 0.64 \\ \hline
professional\_law & 0.3937 & 0.5815 \\ \hline
professional\_psychology & 0.5474 & 0.732 \\ \hline
management & 0.6408 & 0.8252 \\ \hline
public\_relations & 0.5273 & 0.7091 \\ \hline
medical\_genetics & 0.7 & 0.88 \\ \hline
college\_medicine & 0.5549 & 0.7283 \\ \hline
prehistory & 0.608 & 0.7685 \\ \hline
college\_computer\_science & 0.53 & 0.69 \\ \hline
marketing & 0.7479 & 0.906 \\ \hline
high\_school\_psychology & 0.7358 & 0.8936 \\ \hline
conceptual\_physics & 0.6298 & 0.766 \\ \hline
human\_aging & 0.5605 & 0.6951 \\ \hline
professional\_medicine & 0.7132 & 0.8456 \\ \hline
college\_physics & 0.6765 & 0.8039 \\ \hline
machine\_learning & 0.5536 & 0.6786 \\ \hline
moral\_scenarios & 0.3676 & 0.4883 \\ \hline
college\_chemistry & 0.53 & 0.65 \\ \hline
formal\_logic & 0.4206 & 0.5397 \\ \hline
miscellaneous & 0.7893 & 0.908 \\ \hline
virology & 0.4036 & 0.5181 \\ \hline
high\_school\_mathematics & 0.5444 & 0.6556 \\ \hline
anatomy & 0.5778 & 0.6889 \\ \hline
logical\_fallacies & 0.7362 & 0.8466 \\ \hline
world\_religions & 0.7544 & 0.848 \\ \hline
business\_ethics & 0.61 & 0.7 \\ \hline
elementary\_mathematics & 0.7672 & 0.8439 \\ \hline
global\_facts & 0.57 & 0.5 \\ \hline
\caption{Full results for MMLU (57 tasks).} \label{tab:full_mmlu} \\

\end{longtable}
}

{\tiny
\begin{longtable}[l]{|l|c|c|}
\hline
\textbf{Subset} & \textbf{Score (options not in prompt)} & \textbf{Score (options in prompt)} \\ \hline
\endhead

Electronics & 0.302 & 0.5302 \\ \hline
History & 0.5292 & 0.7468 \\ \hline
Psychology & 0.497 & 0.7134 \\ \hline
Design & 0.6683 & 0.8492 \\ \hline
Manage & 0.3613 & 0.5393 \\ \hline
Sociology & 0.4769 & 0.6299 \\ \hline
Math & 0.3671 & 0.4968 \\ \hline
Chemistry & 0.4326 & 0.5603 \\ \hline
Diagnostics\_and\_Laboratory\_Medicine & 0.3229 & 0.4427 \\ \hline
Agriculture & 0.4146 & 0.5316 \\ \hline
Art\_Theory & 0.6419 & 0.7533 \\ \hline
Biology & 0.474 & 0.5836 \\ \hline
Physics & 0.4051 & 0.5118 \\ \hline
Materials & 0.3627 & 0.4481 \\ \hline
Clinical\_Medicine & 0.5847 & 0.6695 \\ \hline
Geography & 0.3811 & 0.4656 \\ \hline
Marketing & 0.625 & 0.7067 \\ \hline
Basic\_Medical\_Science & 0.5347 & 0.6152 \\ \hline
Mechanical\_Engineering & 0.3296 & 0.4081 \\ \hline
Pharmacy & 0.5418 & 0.6058 \\ \hline
Energy\_and\_Power & 0.3764 & 0.4245 \\ \hline
Economics & 0.5274 & 0.5719 \\ \hline
Art & 0.7395 & 0.7816 \\ \hline
Public\_Health & 0.592 & 0.6315 \\ \hline
Literature & 0.8239 & 0.8551 \\ \hline
Computer\_Science & 0.4803 & 0.5046 \\ \hline
Music & 0.3444 & 0.365 \\ \hline
Architecture\_and\_Engineering & 0.3754 & 0.3483 \\ \hline
Finance & 0.5655 & 0.5329 \\ \hline
Accounting & 0.5486 & 0.495 \\ \hline
Finance & 0.5655 & 0.5329 \\ \hline
Accounting & 0.5486 & 0.495 \\ \hline
\caption{Full results for MMMU (30 tasks).} \label{tab:mmmu_scores} 
\end{longtable}
}

{\tiny
\begin{longtable}[l]{|l|l|c|}
\hline
\textbf{Suite} & \textbf{Dataset} & \textbf{Score} \\ \hline
\endhead
image & EDIS & 0.98 \\ \hline
image & Visual7W-Pointing & 0.972 \\ \hline
image & DocVQA & 0.965 \\ \hline
image & RefCOCO & 0.969 \\ \hline
image & RefCOCO-Matching & 0.942 \\ \hline
image & ImageNet-R & 0.943 \\ \hline
image & TextVQA & 0.913 \\ \hline
image & WebQA & 0.904 \\ \hline
image & InfographicsVQA & 0.882 \\ \hline
image & VisDial & 0.874 \\ \hline
image & ObjectNet & 0.857 \\ \hline
image & VOC2007 & 0.855 \\ \hline
image & ImageNet-1K & 0.85 \\ \hline
image & Wiki-SS-NQ & 0.841 \\ \hline
image & VisualNews\_i2t & 0.856 \\ \hline
image & ChartQA & 0.837 \\ \hline
image & OVEN & 0.836 \\ \hline
image & SUN397 & 0.816 \\ \hline
image & VisualNews\_t2i & 0.839 \\ \hline
image & N24News & 0.816 \\ \hline
image & HatefulMemes & 0.794 \\ \hline
image & MSCOCO\_t2i & 0.801 \\ \hline
image & MSCOCO\_i2t & 0.78 \\ \hline
image & ScienceQA & 0.745 \\ \hline
image & OK-VQA & 0.757 \\ \hline
image & GQA & 0.756 \\ \hline
image & ImageNet-A & 0.721 \\ \hline
image & MSCOCO & 0.732 \\ \hline
image & CIRR & 0.685 \\ \hline
image & VizWiz & 0.674 \\ \hline
image & A-OKVQA & 0.665 \\ \hline
image & NIGHTS & 0.672 \\ \hline
image & Visual7W & 0.592 \\ \hline
image & MCMR & 0.6685 \\ \hline
image & Place365 & 0.525 \\ \hline
image & Country211 & 0.477 \\ \hline
image & FashionIQ & 0.427 \\ \hline
video & UCF101 & 0.941 \\ \hline
video & ActivityNetQA & 0.799 \\ \hline
video & SmthSmthV2 & 0.8 \\ \hline
video & NExTQA & 0.7477 \\ \hline
video & HMDB51 & 0.716 \\ \hline
video & K700 & 0.718 \\ \hline
video & MSVD & 0.7239 \\ \hline
video & QVHighlight & 0.7156 \\ \hline
video & DiDeMo & 0.5916 \\ \hline
video & MVBench & 0.5533 \\ \hline
video & Video-MME & 0.5485 \\ \hline
video & MSR-VTT & 0.526 \\ \hline
video & EgoSchema & 0.522 \\ \hline
video & VATEX & 0.5013 \\ \hline
video & MomentSeeker & 0.4744 \\ \hline
video & Breakfast & 0.4665 \\ \hline
video & YouCook2 & 0.2693 \\ \hline
video & Charades-STA & 0.2682 \\ \hline
visdoc & ViDoSeek-doc & 0.9974 \\ \hline
visdoc & ViDoRe\_syntheticDocQA\_artificial\_intelligence & 0.94 \\ \hline
visdoc & VisRAG\_SlideVQA & 0.9514 \\ \hline
visdoc & ViDoRe\_tabfquad & 0.8964 \\ \hline
visdoc & ViDoRe\_syntheticDocQA\_energy & 0.88 \\ \hline
visdoc & ViDoRe\_syntheticDocQA\_healthcare\_industry & 0.9 \\ \hline
visdoc & VisRAG\_InfoVQA & 0.883 \\ \hline
visdoc & ViDoRe\_infovqa & 0.8583 \\ \hline
visdoc & ViDoRe\_syntheticDocQA\_government\_reports & 0.84 \\ \hline
visdoc & ViDoRe\_arxivqa & 0.842 \\ \hline
visdoc & VisRAG\_ArxivQA & 0.8333 \\ \hline
visdoc & VisRAG\_MP-DocVQA & 0.8359 \\ \hline
visdoc & VisRAG\_ChartQA & 0.8571 \\ \hline
visdoc & MMLongBench-doc & 0.7554 \\ \hline
visdoc & ViDoSeek-page & 0.7145 \\ \hline
visdoc & ViDoRe\_shiftproject & 0.65 \\ \hline
visdoc & ViDoRe\_biomedical\_lectures\_v2\_multilingual & 0.6234 \\ \hline
visdoc & ViDoRe\_esg\_reports\_v2\_multilingual & 0.6053 \\ \hline
visdoc & VisRAG\_PlotQA & 0.5817 \\ \hline
visdoc & ViDoRe\_esg\_reports\_human\_labeled\_v2 & 0.6731 \\ \hline
visdoc & ViDoRe\_tatdqa & 0.5765 \\ \hline
visdoc & ViDoRe\_economics\_reports\_v2\_multilingual & 0.6422 \\ \hline
visdoc & MMLongBench-page & 0.5215 \\ \hline
visdoc & ViDoRe\_docvqa & 0.5366 \\ \hline
\caption{Full results on MMEB-V3 image, video and visdoc (79 tasks).} \label{tab:mmeb_full} \\
\end{longtable}
}

{\tiny
\begin{longtable}[l]{|l|c|}

\hline
\textbf{Task} & \textbf{NDCG@10} \\ \hline
\endhead

Touche2020 * & 0.4609 \\ \hline
NQ & 0.8311 \\ \hline
QuoraRetrieval & 0.949 \\ \hline
ClimateFEVER & 0.2869 \\ \hline
FiQA2018 & 0.7123 \\ \hline
MSMARCO & 0.5819 \\ \hline
DBPedia * & 0.6235 \\ \hline
ArguAna & 0.6537 \\ \hline
FEVER & 0.9534 \\ \hline
NFCorpus & 0.3789 \\ \hline
HotpotQA * & 0.8943 \\ \hline
SCIDOCS * & 0.4885 \\ \hline
SciFact & 0.8075 \\ \hline
\caption{Full results on Nano-BEIR (13 tasks).} \label{tab:full_beir} \\
\end{longtable}
}

\end{document}